\documentclass[final,3p,times,twocolumn]{elsarticle}
\usepackage{caption}
\usepackage{courier}
\usepackage{subfigure}
\usepackage{graphicx}
\graphicspath{{./figs/}}

\usepackage{multirow}
\usepackage{booktabs}
\usepackage{array}

\usepackage[bookmarks=false]{hyperref}
\hypersetup{
    colorlinks = true,
    citecolor  = blue,
    linkcolor  = blue,
    urlcolor   = blue,
}

\newcommand{\add}[1]{\textcolor{black}{#1}}
\usepackage{natbib}

\usepackage{amssymb}
\usepackage{amsmath}
\usepackage{bm}
\usepackage{cleveref}

\journal{Neurocomputing}

\begin{document}

\begin{frontmatter}

%% Title, authors and addresses

%% use the tnoteref command within \title for footnotes;
%% use the tnotetext command for the associated footnote;
%% use the fnref command within \author or \address for footnotes;
%% use the fntext command for the associated footnote;
%% use the corref command within \author for corresponding author footnotes;
%% use the cortext command for the associated footnote;
%% use the ead command for the email address,
%% and the form \ead[url] for the home page:
%% \title{Title\tnoteref{label1}}
%% \tnotetext[label1]{}
%% \author{Name\corref{cor1}\fnref{label2}}
%% \ead{email address}
%% \ead[url]{home page}
%% \fntext[label2]{}
%% \cortext[cor1]{}
%% \address{Address\fnref{label3}}
%% \fntext[label3]{}

%\title{An Overview on Recent Advances for Convolutional Neural Network Acceleration}
\title{\add{\textbf{Recent Advances in Convolutional Neural Network Acceleration}} }

%% use optional labels to link authors explicitly to addresses:
%% \author[label1,label2]{}
%% \address[label1]{}
%% \address[label2]{}

%% Group authors per affiliation:
%\author{Qianru Zhang\fnref{myfootnote}}

%% or include affiliations in footnotes:
\author[mymainaddress]{Qianru Zhang}
\author[mymainaddress]{Meng Zhang\corref{mycorrespondingauthor}}
\ead{zmeng@seu.edu.cn}
\cortext[mycorrespondingauthor]{Corresponding author}
\author[mysecondaryaddress]{Tinghuan Chen}
\author[mymainaddress]{Zhifei Sun}
\author[mysecondaryaddress]{Yuzhe Ma}
\author[mysecondaryaddress]{Bei Yu}

\address[mymainaddress]{National ASIC System Engineering Technology Research Center, Southeast University, Nanjing, China}
\address[mysecondaryaddress]{Department of Computer Science \& Engineering, The Chinese University of Hong Kong, Hong Kong}

\begin{abstract}
    In recent years, convolutional neural networks (CNNs) have shown great performance in various fields such as image classification, pattern recognition, and multi-media compression.
    Two of the feature properties, local connectivity and weight sharing, can reduce the number of parameters and increase processing speed during training and \add{inference}.
    However, as the dimension of data becomes higher and the CNN architecture becomes more complicated, the end-to-end approach or the combined manner of CNN is computationally intensive,
    which becomes limitation to CNN's further implementation.
    Therefore, it is necessary and urgent to implement CNN in a faster way.
    In this paper, we first summarize the acceleration methods that contribute to but not limited to CNN by reviewing a broad variety of research papers.
    We propose a taxonomy in terms of three levels, i.e.~structure level, algorithm level, and implementation level, for acceleration methods.
    We also analyze the acceleration methods in terms of CNN architecture compression, algorithm optimization, and hardware-based improvement.
    At last, we give a discussion on different perspectives of these acceleration and optimization methods within each level.
    The discussion shows that the methods in each level still have large exploration space.
    By incorporating such a wide range of disciplines, we expect to provide a comprehensive reference for researchers who are interested in CNN acceleration.
\end{abstract}


\end{frontmatter}

%% \linenumbers

%% main text
\section{Introduction}
\label{intro}
Convolutional neural network (CNN) architectures have been around for over two decades.
Compared with other neural network models such as multiple layer perceptron (MLP),
CNN is designed to take multiple arrays as input and then process the input using convolution operator within a local field by mimicking eyes perceiving images.
Therefore, it shows excellent performance in solving computer vision problems such as image classification,
recognition and understanding \cite{yu2017convolutional,farabet2013learning,wang2016parallel}.
It is also effective for a wide range of fields such as speech recognition that requires correlated speech spectral representations \cite{qin2016empirical},
VLSI physical design \cite{yu2015machine}, multi-media compression \cite{iclr2017_theis_lossy} comparing with the traditional DCT transformation and compressive sensing methods \cite{zhang2016image,chen2016image}, and cancer detection from a series of condition changing images \cite{jothi2017survey}.
Moreover, many top players have been in a fever to play Go match with alphaGo recently, which has CNN implemented. 

However, in order to receive good performance of prediction and accomplish more difficult goals, CNN architecture becomes deeper and more complicated.
At the same time, more pixels are packed into one image thanks to high resolution acquisition devices.
As a result, CNN training and \add{inference} are very computationally expensive and become limited for implementation due to its slow speed.
Although acceleration and optimization for CNN have been explored since it was brought up, recently this seems to be keener as it has such good industrial impact. 

\begin{figure*}[tb!]
    \centering
    \includegraphics[width=0.65\textwidth]{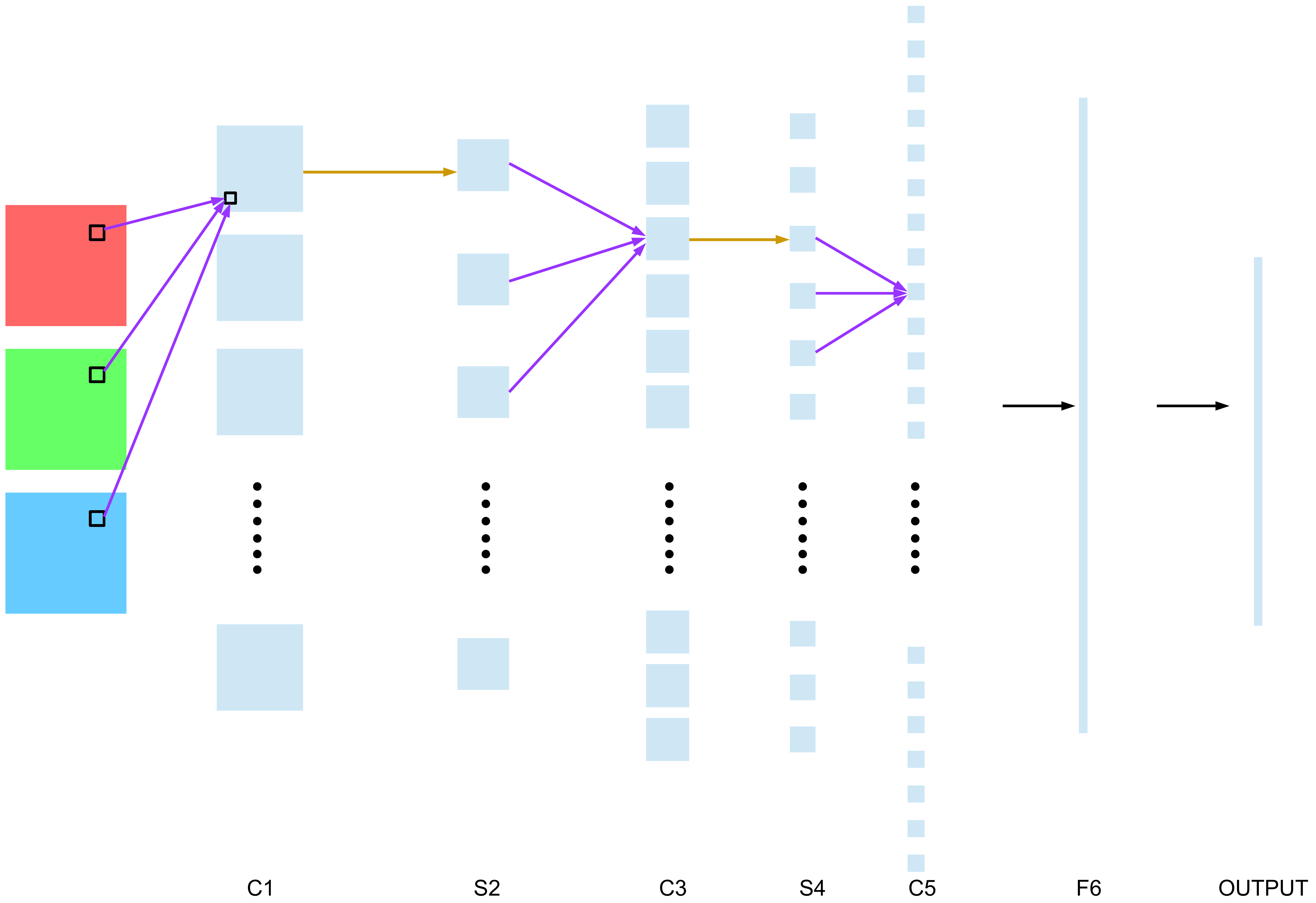}
    \caption{Illustration of LeNet-5.}
    \label{fig:lecunCNN}     
\end{figure*}

Some companies have unveiled accelerators for deep learning inference that can be extensively used for CNN. Google's second generation \add{Tensor Processing Unit} (TPU) is designed for TensorFlow framework that has the increased performance of 92TFLOPS in peak and on-chip memory of 28MiB. It not only supports integers but also floating point calculations, which makes it more powerful in deep learning training \cite{jouppi2017datacenter}. 
NVIDIA launches an open source project called NVIDIA Deep Learning Accelerator (NVDLA) along with an open license that is ready for people who are interested in data intensive automotive products. It includes Verilog and C-model for the chip, Linux drivers, test suites, and kernel and user based software with development tools \cite{nvdla}. 
Intel's Nervana Neural Network Processor (NNP) has been announced recently for dealing with neural network matrix multiplications and convolutions. The memory architecture is designed for efficient operations and data movements without cache hierarchy, which improves the training performance of deep learning \cite{nnp}. 

In this paper, we review many recent works and summarize acceleration methods not only in structure level and algorithm level, but also in implementation level. This paper differs from other deep neural network review papers in three aspects. \textbf{1) Topic:} Some review papers summarize the relevant work regarding deep neural networks in different applications such as computer vision, natural language processing and acoustic signal processing, among which CNN is only a component \cite{guo2016deep,ghayoumi2017quick,carrio2017review,zhang2015deep,singh2017machine,ling2015deep}. They make systematic introduction for various kinds of neural networks that are fit for specific applications and provide a guide for people who want to implement deep neural networks in their fields. However, few of them mention acceleration methods, while our paper focuses on CNN and its acceleration methods. \textbf{2) Time:} Recent deep learning review papers are mostly historical \cite{schmidhuber2015deep, lecun2015deep, du2016overview}. They usually trace back the origins through over fifteen years to form a big picture of the neural network development, which is very inspiring to think over the origins. Our paper focuses on researches recently when hardware becomes limited and efficiency becomes the priority. \textbf{3) Taxonomy:}  There are no reviews that incorporate hardware into algorithms since they are different disciplines. In this paper, we talk about the acceleration methods in three levels, because they are interwoven and highly dependent.

This survey paper is organized as following.
In Section~\ref{sec:struct}, an overview of modern CNN structure is given with different typical layers that the improvement is focused on.
In Section~\ref{sec:taxonomy} we present our taxonomy for recent CNN acceleration methods followed by the overview in three categories, including CNN compression in Section~\ref{sec:compress}, algorithm optimization in Section~\ref{sec:opt}, and hardware-oriented acceleration in Section~\ref{sec:acce}.
After that, in Section~\ref{sec:discuss} a discussion is given on these methods from different perspectives.
Finally, Section~\ref{sec:conclu} concludes this paper with some future challenges.

\begin{table*}[!htb]
    %{{{
    \centering
    \begin{tabular}{p{1.5cm}p{1.7cm}p{2.9cm}p{2.7cm}p{2.1cm}p{2.6cm}}
        \toprule
        Model & Layer Size & Configuration & Feature & Parameter Size & Application \\\hline\hline
        LeNet \cite{pieee1998_LeCun_lenet} & 7 layers & \add{3C-2S-1F-RBF output layer} &  & 60,000 & Document recognition \\\hline
        AlexNet \cite{nips2012_krizhevsky_alexnet} & 8 layers & \add{5C-3S-3F} & \add{Local response normalization} & 60,000,000 & Image classification \\\hline
        NIN \cite{lin2013network} & - & 3mlpconv-\add{global average pooling} \add{(S can be added in between the mlpconv)} & mlpconv layer: 1C-3MLP; \add{global average pooling} & - & Image classification \\\hline
        VGG \cite{iclr2015_simonyan_vgg} & 11-19 layers & VGG-16: \add{13C-5S-3F} & \add{Increased depth with stacked $3 \times 3$ kernels} & 133,000,000 to 144,000,000 & Image classification and localization \\\hline
        ResNet \cite{cvpr2016_he_resnet} & Can be very deep (152 layers) & ResNet-152: \add{151C-2S-1F} & Residual module & ResNet-20: 270,000; ResNet-1202: 19,400,000 & Image classification, object detection \\\hline
        GoogLeNet \cite{cvpr2015_szegedy_googlenet} & 22 layers & 3C-\add{9}Inception-5S-1\add{F} & Inception module & 6,797,700 & Image classification, object detection \\\hline
        Xception \cite{chollet2017xception} & 37 layers & 36C-\add{5S}-1\add{F} & \add{Depth-wise separable convolutions} & 22,855,952 & Image classification \\
        \bottomrule
    \end{tabular}
    \captionsetup{justification=centering}
    \caption[caption]{\add{CNN model summary. \\C: convolutional layer, S: subsampling layer, F: fully-connected layer}}
    %}}}
    \label{tab:model}
\end{table*}

\section{Convolutional Neural Network}
\label{sec:struct}

The modern convolutional neural networks proposed by LeCun \cite{pieee1998_LeCun_lenet} is a 7-layer (excluding the input layer) LeNet-5 structure.
It has the following structure C1, S2, C3, S4, C5, F6, OUTPUT as shown in Figure~\ref{fig:lecunCNN},
where C indicates convolutional layer, S indicates subsampling layer, and F indicates fully-connected layer.
There are many modifications regarding the structure of CNNs in order to handle more complicated datasets and problems, such as AlexNet (8 layers) \cite{nips2012_krizhevsky_alexnet},
GoogLeNet (22 layers) \cite{cvpr2015_szegedy_googlenet}, VGG-16 (16 layers) \cite{iclr2015_simonyan_vgg}, and ResNet (152 layers) \cite{cvpr2016_he_resnet}. \add{Table \ref{tab:model} summarizes the state-of-the-art CNNs. In this table, \add{Feature column summarizes the most important parts in each model \cite{aloysius2017review,al2017review,rawat2017deep}. Application column provides the fields that the methods were proposed for the first time. Fully-connected layer is followed by the Softmax layer except for LeNet and NIN.} As we can see from the table, the number of parameters in modern CNNs is large, which usually takes a long time for training and for inference. Plus, higher dimensional input, large number of parameters, and complex CNN configuration challenge hardware in terms of processing element efficiency, memory bandwidth, off-chip memory, communication and so on.}

%\begin{table*}[!htb]
%{{{
%   \centering
%   \begin{tabular}{p{1.5cm}|p{2cm}|p{2cm}|p{2cm}|p{2.5cm}|p{2.5cm}}
%       \hline
%               Model & Layer Size & Configuration & Feature & Parameter Size & Application \\
%               \hline
%       \hline
%               LeNet \cite{pieee1998_LeCun_lenet} & 7 layers & 3C-2S-1F &  & 60,000 & \multirow{7}{2.3cm}{Document recognition, computer vision, speech recognition, fraud detection, high-resolution remote sensing, etc.} \\
%               \cline{1-5}
%               AlexNet \cite{nips2012_krizhevsky_alexnet} & 8 layers & 5C-3F (pooling at C1, C2, C5) & Training with 2 GPUs & 60,000,000 &~\\
%       \cline{1-5}
%       NIN \cite{lin2013network} & - & 3mlpconv-1S & mlpconv layer: 1C-3MLP & - &~\\
%       \cline{1-5}
%       VGG \cite{iclr2015_simonyan_vgg} & 11-19 layers & VGG-16: 13C-5S-3F-1softmax & Very small filter kernel & 133,000,000 to 144,000,000 &~\\
%       \cline{1-5}
%       ResNet \cite{cvpr2016_he_resnet} & Can be very deep (152 layers) & ResNet-152: 151C-1F & Residual module & ResNet-20: 270,000; ResNet-1202:19,400,000 &~ \\
%       \cline{1-5}
%       GoogLeNet \cite{cvpr2015_szegedy_googlenet} & 22 layers & 3C-18Inception-5S-1linear & Inception module & 6,797,700 &~ \\
%       \cline{1-5}
%       Xception \cite{chollet2017xception} & 37 layers & 36C-1fc & Separable convolution layer & 22,855,952 &~ \\
%               \hline
%   \end{tabular}
%   \caption{\add{CNN model summary}}
%   \label{tab:model}
%}}}
%\end{table*}

Among these different structures, they share four key features including weight sharing, local connection, pooling, and the use of many layers \cite{lecun2015deep}. There are some commonly used layers such as convolutional layers, subsampling layers (pooling layers), and fully-connected layers. Usually, there is a convolutional layer after the input. The convolutional layer is often followed by a subsampling layer. This combination repeats several times to increase the depth of CNN. The fully-connected layers are designed as the last few layers in order to map from extracted features to labels. These four layers are introduced as follows.

\textbf{a) Input Layer:} In CNNs, input layers usually take multiple arrays and are often size-fixed. Comparing to ordinary fully-connected neural networks, the CNN input do not need size-normalization and centralization, because CNN enjoys the characteristic of translation invariance \cite{bishop2006pattern}.  

\textbf{b) Convolutional Layer:} As a key feature layer that makes CNNs different from other ordinary neural networks, neuron units of convolutional layers are first computed by convolution operation over small local patches of input, and then followed by activation functions (tanh, sigmoid, ReLU, etc.), and form a 2D feature map (3D feature map channel). In general, we have that 
\add{\begin{align}
\label{eq:conv1}
    \mathbf{Z}_{j} &= \sum_{i} \mathbf{X}_{i} \ast \mathbf{K}_{ij} + \mathbf{B}_{j}, \\
\label{eq:conv2}
    \mathbf{A}_{j} &= f(\mathbf{Z}_{j}),
\end{align}}
%\begin{align}
%    z_{j} &= \sum_{i} x_{i} \ast k_{ij} + b_{j}, \\
%    a_{j} &= f(z_{j}),
%\end{align}
where \add{$\mathbf{Z}_{j}$} represents the output from the convolution operation, \add{$\mathbf{X}_{i}$} denotes the input to the convolutional layer, \add{$\mathbf{K}_{ij}$} is the convolution kernel, and \add{$\mathbf{B}_{j}$} is the additive bias. In the following equation, \add{$\mathbf{A}_{j}$} is the output feature map of the convolutional layer and $f(\cdot)$ is an activation function. 

\begin{figure*}[!htb]
    %{{{
    \centering
    \subfigure[Sigmoid]{
        \label{fig:act_a}
        \fbox{\includegraphics[width=0.25\textwidth]{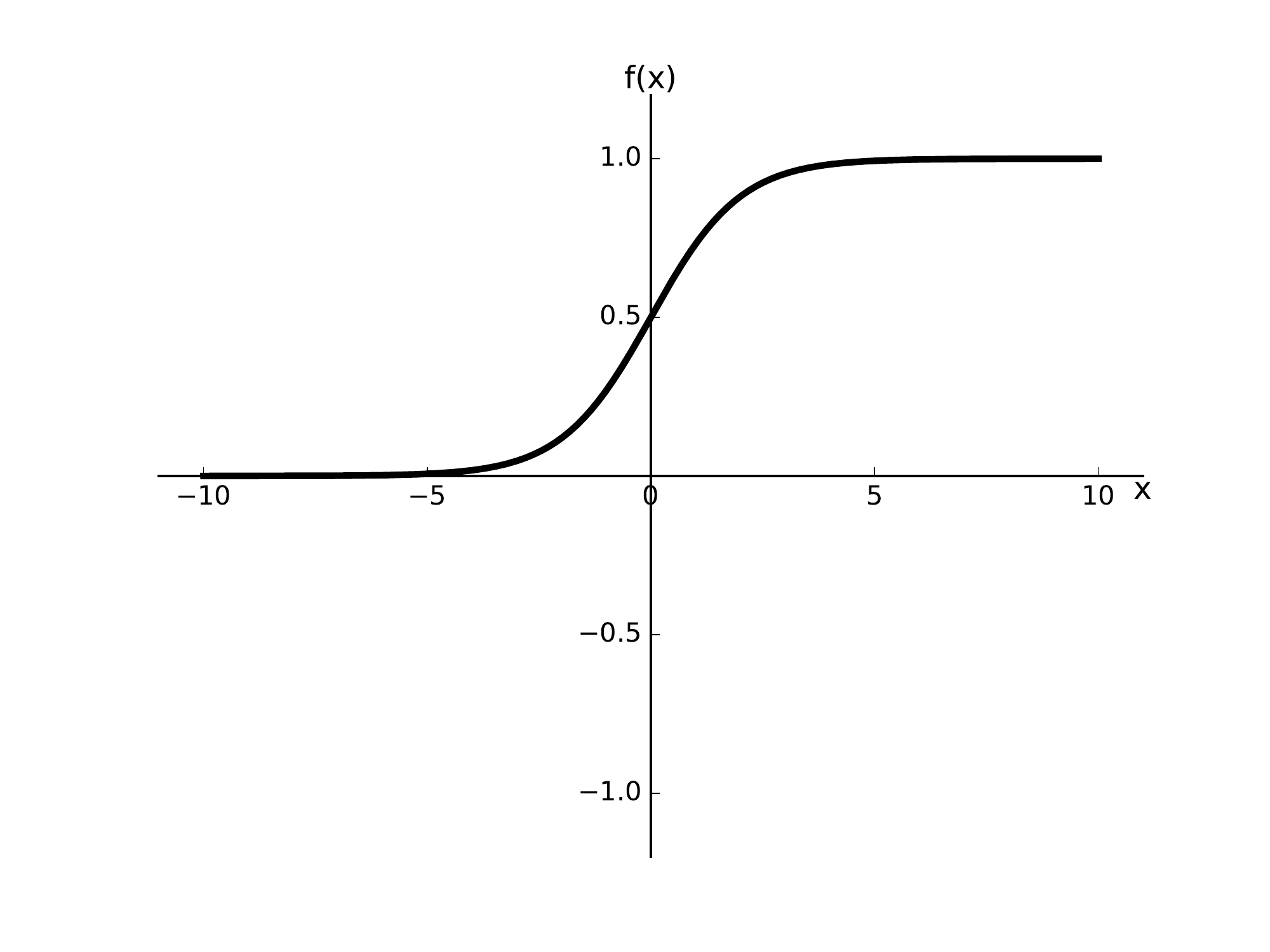}} }
    \subfigure[Tanh]{
        \label{fig:act_b}
        \fbox{\includegraphics[width=0.25\textwidth]{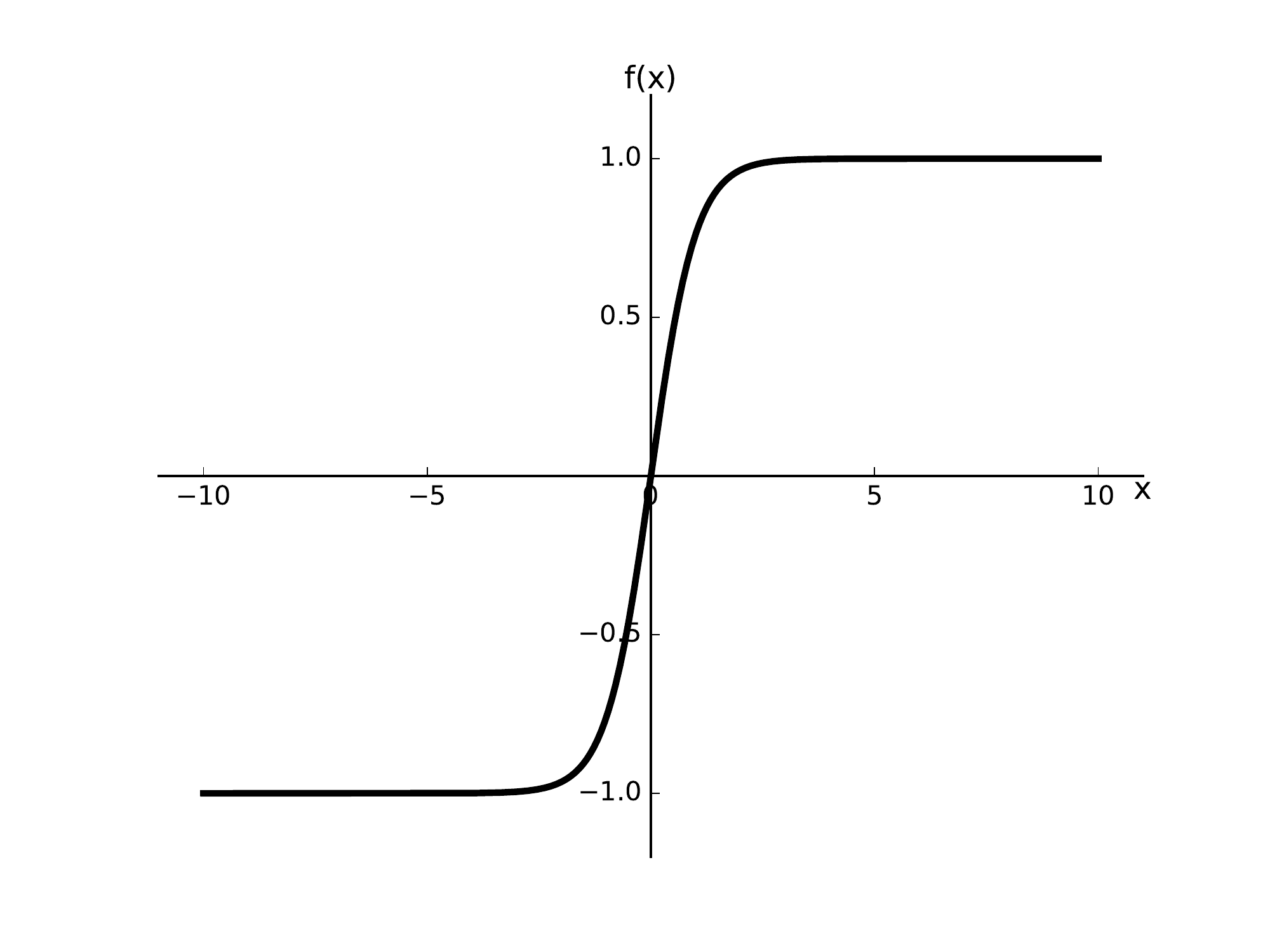}} }
    \subfigure[ReLU]{
        \label{fig:act_c}
        \fbox{\includegraphics[width=0.25\textwidth]{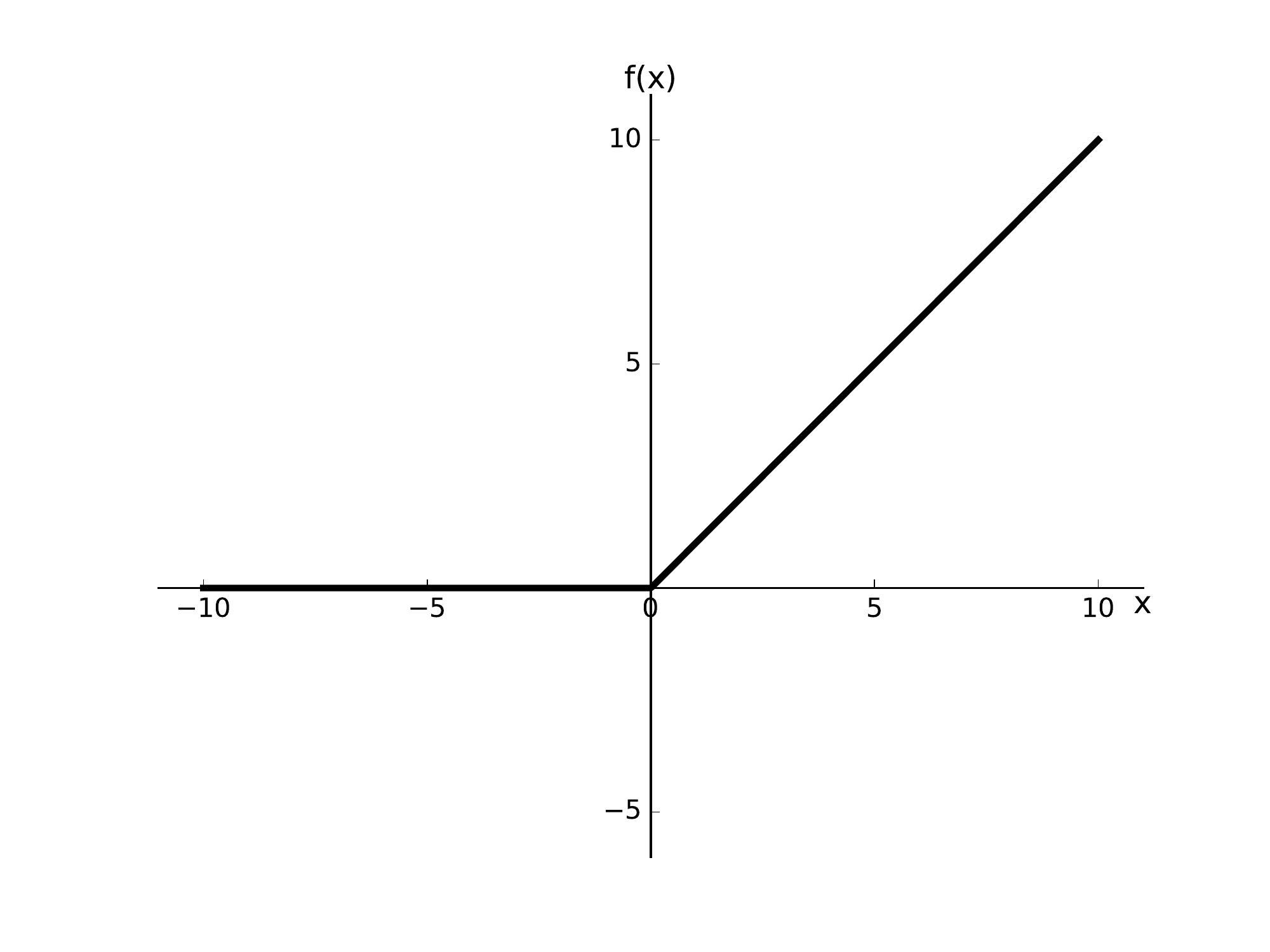}} }
    \subfigure[LeakyReLU / PReLU / RReLU]{
        \label{fig:act_d}
        \fbox{\includegraphics[width=0.25\textwidth]{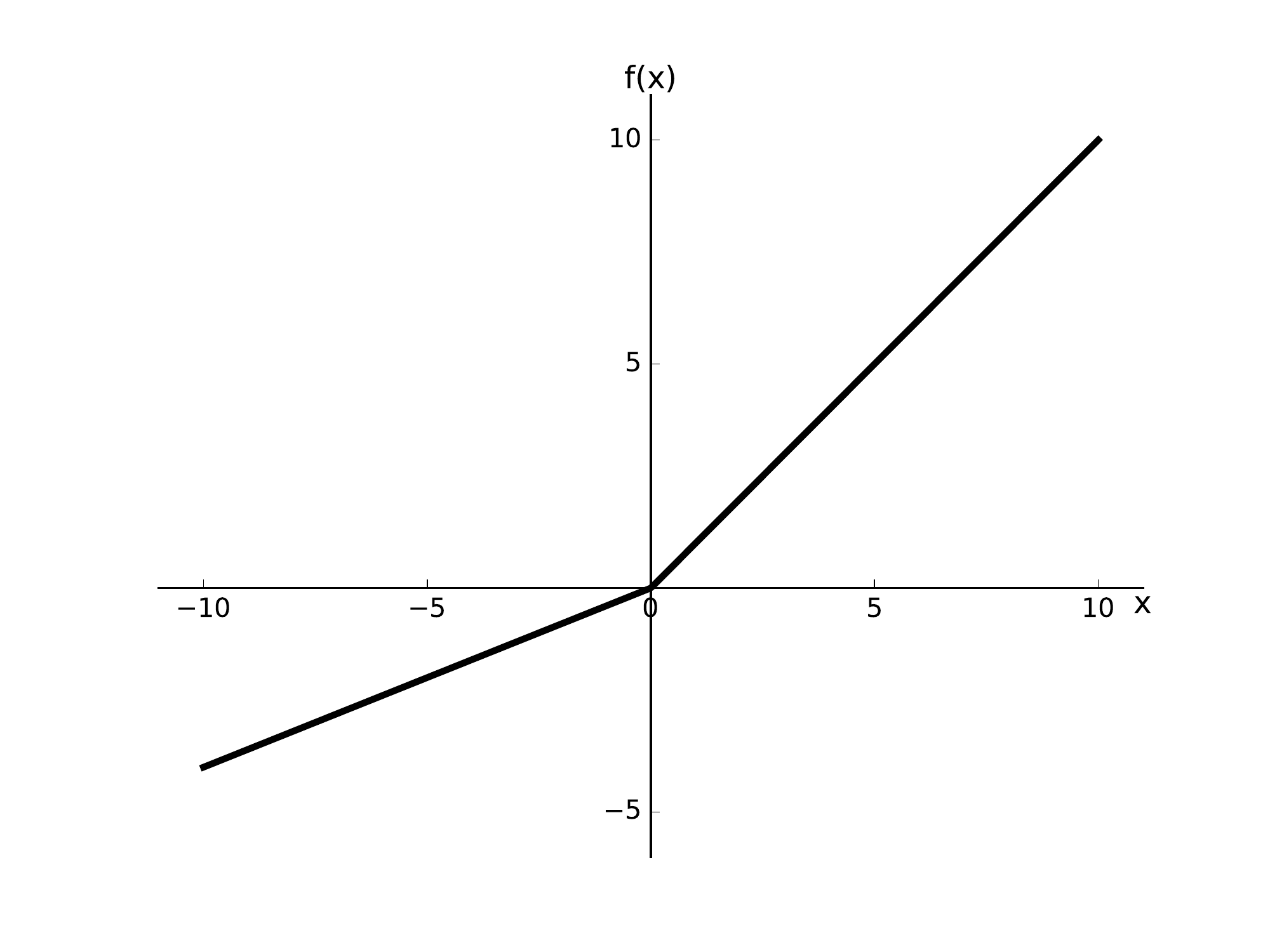}} }
    \subfigure[ELU]{
        \label{fig:act_e}
        \fbox{\includegraphics[width=0.25\textwidth]{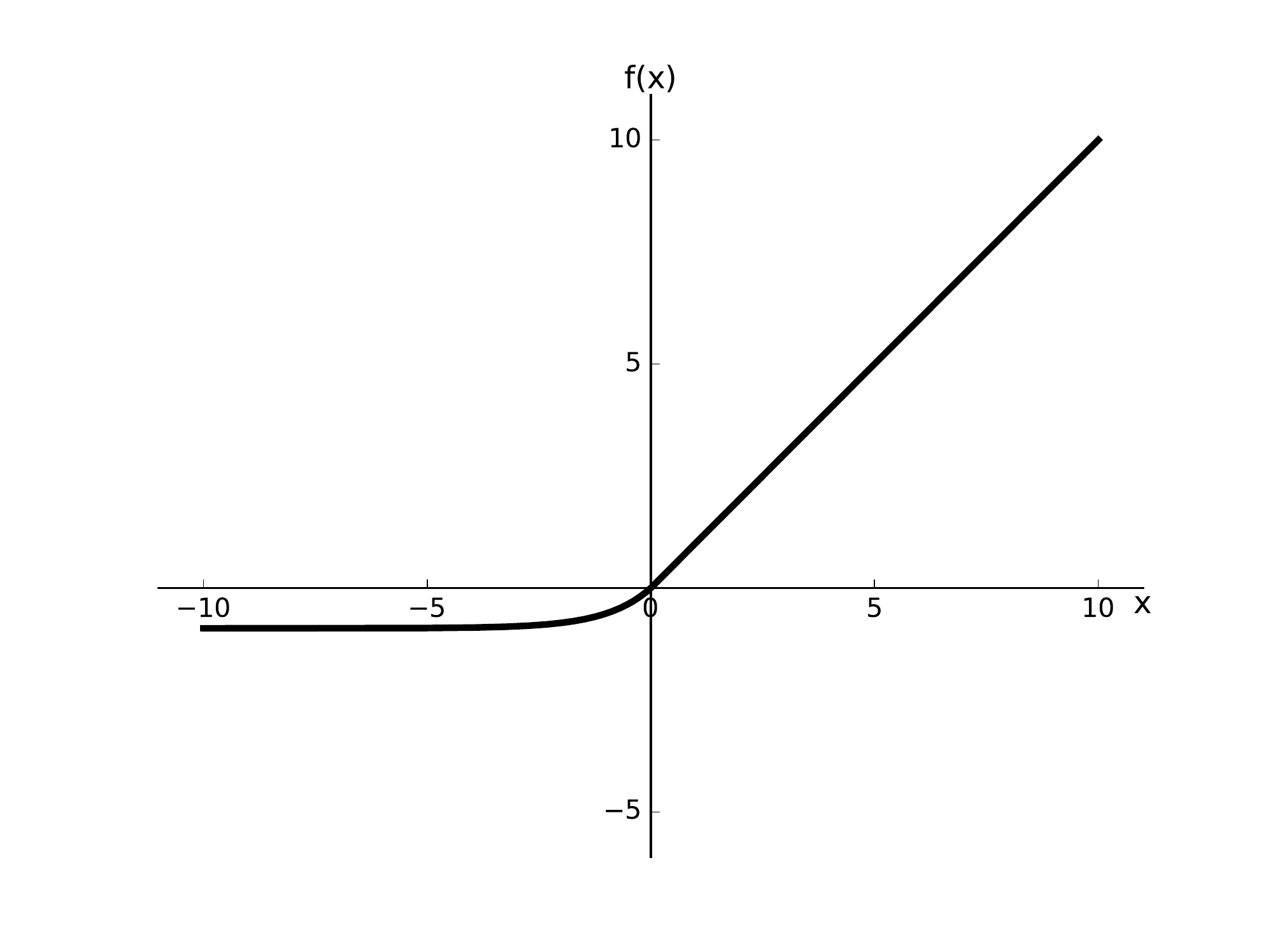}} }
    \caption{Activation function plot.}
    %}}}
    \label{fig:act}
\end{figure*}

\begin{table*}[!htb]
    %{{{
    \centering
    \begin{tabular}{p{1.5cm}cccp{0.5cm}}
    \toprule
    Function & Saturation    & Definition                                                                              & Parameter $\alpha$                 & Plot \\ \hline \hline
    Sigmoid  & Saturated     & \(\displaystyle f(x)=1/(1+e^{-x}) \)                                              & -                                  & (a)  \\ \hline
    Tanh     & Saturated     & \(\displaystyle f(x)=2/(1+e^{-2x}) -1 \)                                          & -                                  & (b)  \\ \hline
    ReLU     & Non-saturated & \(\displaystyle f(x)=\begin{cases} x & x\geq 0 \\ 0 & x<0 \end{cases} \)                & -                                  & (c)  \\ \hline
    LeakyReLU& Non-saturated & \(\displaystyle f(x)=\begin{cases} x & x\geq 0 \\ \alpha x & x<0 \end{cases} \)         & $\alpha \in (0,1)$                 & (d)  \\ \hline
    PReLU    & Non-saturated & \(\displaystyle f(x)=\begin{cases} x & x\geq 0 \\ \alpha x & x<0 \end{cases} \)         & $\alpha$ is a learned parameter    & (d)  \\ \hline 
    RReLU    & Non-saturated & \(\displaystyle f(x)=\begin{cases} x & x\geq 0 \\ \alpha x & x<0 \end{cases} \)         & $\alpha \sim$ uniform(a, b)        & (d)  \\ \hline
    ELU      & Non-saturated & \(\displaystyle f(x)=\begin{cases} x & x\geq 0 \\ \alpha (e^{x}-1) & x<0 \end{cases} \) & $\alpha$ is a predefined parameter & (e)  \\
    \bottomrule
    \end{tabular}
    \caption{\add{Activation function summary.}}
    %}}}
    \label{tab:act}
\end{table*}

\add{Activation functions are mathematical operations over the input, which introduces non-linearity into neural networks and help catch non-linear features of the input data. There are various types of activation functions as summarized in Table~\ref{tab:act} and Figure~\ref{fig:act}.} 

\add{Sigmoid and Tanh are called saturated functions. As we can see from their definitions or plots, when the input is very small or very large, the output saturates at 0 or 1 for Sigmoid and -1 or 1 for Tanh. There are two problems with saturation. The gradients at saturated regions are almost zero, which dramatically decreases neurons’ backpropagation and makes it difficult to converge in the training phase. Furthermore, more attention needs to be paid in weight initialization when using saturated activation functions for the neural networks may not learn in the first place. To alleviate saturation problem, many non-saturated activations are proposed such as Rectified Linear Unit (ReLU) \cite{nair2010rectified}, Leaky ReLU \cite{maas2013rectifier}, Parametric ReLU (PReLU) \cite{he2015delving}, Randomized Leaky ReLU (RReLU) \cite{xu2015empirical}, and Exponential Linear Unit (ELU) \cite{clevert2015fast}.}

Convolution plays a very important role in CNN. On one hand, by weight sharing, neurons in the same feature map share the same parameters, which reduces dramatically the total number of parameters. In different spatial location, input may have some same features such as edges, points, angles, etc. Weight sharing makes the CNN less sensitive to location and shifting. On the other hand, since each convolution operation is targeted for a small patch of input, the extracted features remain intrinsic topology of the input that helps recognize patterns. 

\textbf{c) Subsampling Layer (pooling layer):} Convolutional layers are usually followed by subsampling layers to reduce the feature map resolution. \add{The amount of parameters and computation are also reduced accordingly.} More formally, 
\begin{equation}
\add{
\mathbf{Z}_{j} = \mathrm{down}(\mathbf{X}_{j}),
}
\end{equation}
where $\mathrm{down}(\cdot)$ represents a subsampling method. 

\add{Maximum operation and average operation are two typical subsampling methods and have been implemented in CNNs. In spite of max pooling and average pooling, some methods that work better in mitigating overfitting problems in CNN are proposed such as Lp pooling \cite{sermanet2012convolutional}, stochastic pooling \cite{zeiler2013stochastic}, and mixed pooling \cite{yu2014mixed}.
He \textit{et al}.~propose a pooling method called spatial pyramids pooling (SPP) that can output a fixed-length feature map and therefore can deal with various input image sizes \cite{he2015spatial}. Spectral pooling is a pooling method to reduce dimensionality in frequency, which preserves more information than spacial domain, and can be implemented in Fast Fourier Transform (FFT) based CNNs \cite{rippel2015spectral}.
While multi-scale orderless pooling proposed by Gong \textit{et al}.~outperforms other methods in highly variable scene matching \cite{gong2014multi}.}

Different from convolution kernels, subsampling kernels are often hand-picked and remain unchanged during training and inference. There are two main reasons for subsampling. One is that by maximizing or averaging over the previous feature map, the size of feature map reduces. The other one is that by subsampling, the output feature map is more robust to distortions and errors of individual neuron units \cite{liu2017survey}. 

\textbf{d) Fully-connected Layer:} After several layers, high-level features are extracted and require mapping to \add{labels}.
In fully-connected layer, neuron units are transformed from 2D into 1D. Each unit in the current layer is connected to all the units in the previous layer such like regular neural networks.
It not only extracts features in a more complex way in order to dig deep for more information, but patterns in different locations are connected as well.

\textbf{e) Output Layer:} As a feed-forward neural network, the output layer neuron units are fixed.
They are usually linked with previous neurons in a fully-connected way and they are the final threshold for predicting. 

In general, CNNs have gained a lot of interest in researching the meaning behind the combination of those different layers.
The advantages brought by the structure of CNNs include reduced number of parameters and translation invariance.

\begin{figure*}[!htb]
    \begin{center}
    \includegraphics[width=0.8\textwidth]{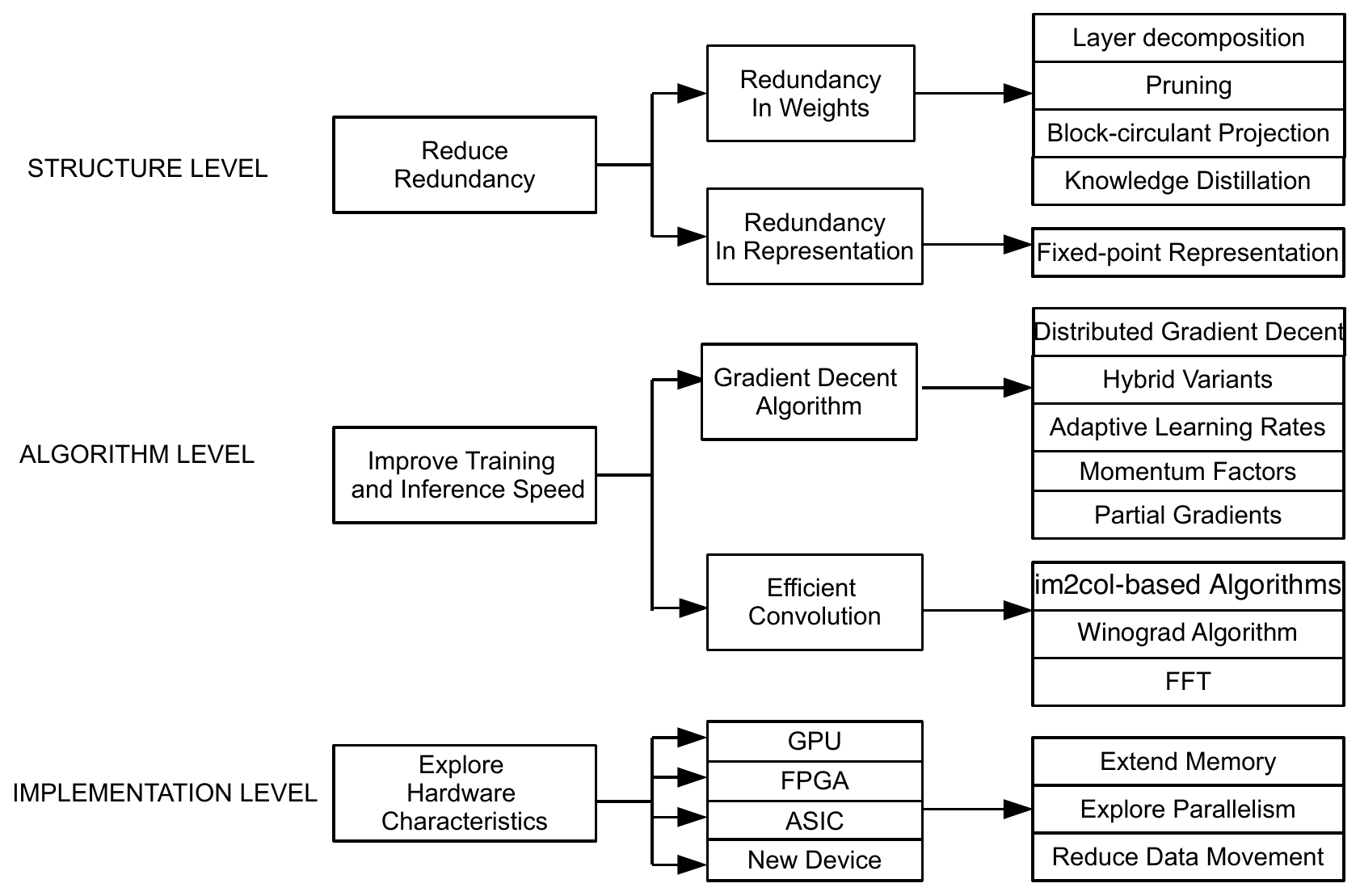}
    \end{center}
    \caption{Taxonomy of CNN acceleration methods.}
    \label{fig:taxonomy}  
\end{figure*}

\section{Acceleration Method Taxonomy}
\label{sec:taxonomy}

Our taxonomy is shown in Figure~\ref{fig:taxonomy}.
The philosophy behind the taxonomy is the order from designing, to training a CNN and finally to implementing it on hardware.
For the CNN structure, there is redundancy in both weights and the number of bits for representation.
For the redundancy in weights, layer decomposition, network pruning, \add{block-circulant projection} and knowledge distillation methods can be applied.
For the redundancy in representation, using fixed-point representation is the mainstream.
Once the structure is decided, CNN adopts training algorithms that are generally used in other neural networks for training process.
The most popular training method is gradient decent based back-propagation algorithm.
By propagating errors back from output to input and by adjusting weights wired in the network, errors can be reduced to an acceptable degree.
The criterion for algorithm optimization is convergence speed with proper stability.
Considering that convolutional layers are computationally intensive, we are also interested in the convolution operation complexity.
Therefore, we also summarize some efficient convolution methods that are adopted in the CNN.
As for the implementation level, the mainstream GPU, FPGA, ASIC are discussed.
Recently, people see a promising future for fast implementation of CNN as neuromorphic engineering develops.
Some new \add{devices are} also presented in this paper.
The acceleration approaches of each level is orthogonal and can be combined with those in other levels.
By researching such a wide range of methods, we expect to provide a general idea on CNN acceleration, especially for deep CNN, from the perspectives of structure, algorithm, and hardware.

\section{Structure Level}
\label{sec:compress}

Many training and inference process can be accelerated by reducing redundancy in network structures. There is redundancy both in weights and in the way how weights are represented. Two perspectives of acceleration methods will be summarized as follows in terms of redundancy in weights and redundancy in representations.

\subsection{Redundancy In Weights}
There is significant redundancy in the parameterization of some neural networks. As Denil \textit{et al}.~and Sainath \textit{et al}.~observe that some weights learned in networks are correlated with each other, they demonstrate that some of the weights can either be predicted or be unnecessary to learn \cite{nips2013_mdenil_redundancyinnn,icassp2013_tsainath_lowrank}. 

\subsubsection{Layer Decomposition}
Low-rank approximation can be adopted to reduce redundancy in weights \cite{jaderberg2014speeding}. For one layer, the input-output relationship can be described by
\begin{equation}
\mathbf{y}=g(\mathbf{x}\cdot\mathbf{W}),
\end{equation}
where $\mathbf{W}$ is the weight matrix with size $m\times n$. $\mathbf{W}$ can be replaced by the product of two full rank matrices $\mathbf{U}\cdot \mathbf{V}$ with size $m\times r$ and $r \times n$ respectively. \add{The number of parameters in $\mathbf{W}$ can be reduced to $1/d$ if the following inequality holds, $d(mr+rn)<mn$.} An efficient low-rank approximation of kernels can be applied in first few convolutional layers of CNN to exploit the linear structure of the over-parameterization within a filter. For example, Denton \textit{et al}.~reduce the computation work for redundancy within kernels. It achieves $2\sim2.5\times$ speedup with less than $1\%$ drop in classification performance for a single convolutional layer. It uses singular value decomposition method to exploit the approximation of kernels with assumptions that the singular values of the kernels decay rapidly so that the size of the kernels can be reduced significantly  \cite{nips2014_edentonylecun_linearstructureconvapprox}. 

\add{Instead of treating kernel filters as different matrices, kernels in one layer can be treated as a 3D tensor with two spatial dimensions and the third dimension representing channels. Lebedev \textit{et al}.~use CP-decomposition for convolutional layers, which achieves $8.5\times$ CPU speedup at the cost of $1\%$ error increase \cite{iclr2015_vlebedev_cpdecomposition}. Tai \textit{et al}.~utilize tensor decomposition to remove the redundancy in the convolution kernels, which achieves twice more efficiency of inference for VGG-16 \cite{tai2015convolutional}. Wang \textit{et al}.~propose to use group sparse tensor decomposition for each convolutional layer, which achieves $6.6\times$ speed-up on PC and $5.91\times$ speed-up on mobile device with less than $1\%$ error rate increase \cite{wang2016accelerating}. Tucker decomposition is also used recently to decompose pre-trained weights with fine-tuning afterwards  \cite{kim2015compression,ding2017compact}.}

Weight matrix decomposition method can not only be applied to convolutional layers, but also fully-connected layers. Applying the low-rank approximation to the fully-connected layer weight can achieve a $30\sim 50\%$ reduction of number of parameters with little loss in accuracy, which is roughly an equivalent reduction in training time \cite{icassp2013_tsainath_lowrank}. In spite of using two full rank matrices 

\begin{equation}
\mathbf{W}=\mathbf{U}\cdot \mathbf{V},
\end{equation}
some works have proposed different decomposition forms 
\begin{equation}
\mathbf{W}=\mathbf{D}_1\cdot \mathbf{H}\cdot \mathbf{P}\cdot \mathbf{D}_2\cdot \mathbf{H}\cdot \mathbf{D}_3,
\end{equation}
with diagonal matrices $\mathbf{D}_{1,2,3}$ and Hadamard matrix $\mathbf{H}$ \cite{le2013fastfood}, and 
\begin{equation}
\mathbf{W}=\mathbf{A}\cdot \mathbf{C}\cdot \mathbf{D}\cdot \mathbf{C}^{-1},
\end{equation}
with diagonal matrices $\mathbf{A}$, $\mathbf{D}$ and DCT matrix $\mathbf{C}$ \cite{iclr2016_mmoczulski_acdclinearlayer}. This method can be used during training the CNN, which is very meaningful. The CNN efficiency can be further improved if the training complexity can be reduced as well. Ioannou \textit{et al}.~propose to learn some basis small filters that can describe the more complex filters from the scratch. By carefully choosing the initialization status, the new method can be used during training \cite{ioannou2015training}. Wen \textit{et al}.~force more weight information into the filters to get more efficient CNNs. With its help, the training process converges faster during the fine-tuning phase. In the experiments, it obtains $2\times$ faster on GPU without accuracy loss \cite{wen2017coordinating}.

The decomposition technique is layer oriented and can be interleaved with other modules such as ReLU modules in CNN. It can also be applied to the structure of neural networks. Rigamonti \textit{et al}.~apply this technique to the general frameworks and reduce the computational complexity by using linear combinations of fewer separable filters \cite{tpami2015_asironi_separablefilters}. This method can be extended for multiple layers (e.g. $>10$) by utilizing low-rank approximation for both weights and input \cite{tpami2016_xzhang_gsvd}. It can achieve $4\times$ speedup with $0.3\%$ error increase for deep network models VGG-16 by focusing on reducing accumulated error across layers using generalized singular value decomposition. 

The methods above can be generalized as layer decomposition for filter weight matrix dimension reduction, while pruning is another method for dimension reduction. 

\subsubsection{Network Pruning}
\label{subsub:prune}
Network pruning originates as a method to reduce the size and over-fitting of a neural network. As neural network implementation on hardware becomes more popular, it is necessary to reduce the limitation such as its intensive computation and large memory bandwidth requirement. Nowadays, pruning is usually adopted as a method to reduce the network size and to increase the network inference speed so that it can be applied in specific hardware such as embedded systems. 

\add{There are many pruning methods in terms of weights, connections, filters, channels, feature maps, and so on.}
Unlike layer decomposition in which computational complexity is reduced through reducing the total size of layers, selected neurons are removed in pruning. For pruning weights, the unimportant connections of weights with magnitudes smaller than a given threshold are dropped. Experiments are taken on NVIDIA TitanX and GTX980 GPUs, which achieves $9\times$ and $13\times$ parameter reduction for AlexNet and VGG-16 models respectively with no loss of accuracy \cite{iclr2016_shan_pruningquantizationhuffman}. Zhou \textit{et al}.~incorporate sparse constraints to decimate the number of neurons during training, which reduces the $70\%$ number of neurons without accuracy sacrifice \cite{zhou2016less}. Besides the method to eliminate least influential neurons, another method is to merge selected and the rest of neurons to maintain diversity in the information. Mariet \textit{et al}.~succeed in merging the qualified neurons with unqualified ones and reduce the network complexity \cite{iclr2016_ZMariet_mergeNeuronPruning}. By trading off the training error with the remaining hidden neurons, they achieve $25\%$ reduction of the original number of parameters with $0.04$ accuracy reduction in MNIST dataset. Channel pruning method is to eliminate lowly active channels, which means filters are applied in fewer number of channels in each layer. Polyak \textit{et al}.~propose a channel-pruning based method Inbound Prune to compress a redundant network. Their experiment is taken on the platform of Samsung Galaxy S6 and it achieves $1.59\times$ speedup \cite{access2015_apolyak_facenetspruning}. Recently, pruning is combined with other acceleration techniques to achieve speedup. For example, Han \textit{et al}.~combine pruning with trained quantization and Huffman coding to deep compress the neural networks in three steps. \add{It achieves $3\times$ layer-wise speedup on fully-connected layer over benchmark on CPU \cite{iclr2016_shan_pruningquantizationhuffman}.}

\add{Some of these pruning methods result in structured sparsity, while others cause unstructured sparsity such as weight-based pruning.
Many techniques are proposed to deal with problems of unstructured sparsity being unfriendly to hardware.
Wen \textit{et al}.~propose a method called Structured Sparsity Learning (SSL) for regularizing compressed structures of deep CNNs and speeding up convolutional computation by group Lasso regularization and locality optimization respectively.
It improves convolutional layer computation speed by 5.1$\times$ and 3.1$\times$ over CPU and GPU \cite{wen2016learning}.
He \textit{el al}.~propose a channel pruning method by iteratively reducing redundant channels through solving LASSO and reconstructing the outputs with linear least squares.
It achieves 5$\times$ speed increase in VGG-16 and 2$\times$ speedup in ResNet/Xception \cite{he2017channel}.
Liu \textit{et al}.~also impose channel-based pruning.
They use L1 regularization and achieve 20$\times$ reduction in model size and 5$\times$ reduction in computing operations for VGG model \cite{liu2017learning}.
Li \textit{et al}.~prune whole filters as well as their related feature maps and reduce inference cost of VGG-16 by $34\%$ and ResNet-110 by $38\%$ \cite{li2016pruning}.
Their method uses sum of filter's absolute values as a measurement of filter importance, which is filter-based and avoids sparse connectivity.
Based on Taylor expansion of cost function between pruning and non-pruning situations,
Molchanov \textit{et al}.~reduce feature maps from convolutional layers and implement the iterative pruning method in transfer learning setting \cite{vedaldi2015matconvnet}.
ASIC based methods dealing with irregular sparsity are proposed as well and will be discussed in Section \ref{subsec:asic}.
}

\add{\subsubsection{Block-circulant Projection}
\label{subsub:circulant}
A square matrix could be represented by a one-block-circulant matrix, while a non-squared matrix could be represented by block-circulant matrix. Block-circulant matrix is one of the structured matrices that is usually used in paradigms such as dimension reduction \cite{oymak2017near}, since it can represent an unstructured matrix with a vector. A one-block-circulant matrix is defined as}
\begin{equation}
    \add{
	\textbf{R} = circ(\textbf{r})~:=
	\begin{bmatrix}
	    r_0     & r_{d-1} & \cdots  & r_2     & r_1\\
    	    r_1     & r_0       & r_{d-1} & \cdots & r_2\\
	    \vdots & r_1       & r_0      & \ddots & \vdots\\ 
	    r_{d-1}& r_{d-2}  & \cdots & r_1      & r_0
	\end{bmatrix},}
\end{equation}
\add{which can be represented by a vector $\textbf{r} = (r_0, r_1, \ldots, r_{d-1})$. Block-circulant based CNN has been explored nowadays as it has small storage requirements.}

\add{Cheng \textit{et al}.~apply the circulant matrix in the fully connected layer and achieve significant gain in efficiency with little decrease in accuracy \cite{iccv2015_zyang_fastfoodreparameterize}. Yang \textit{et al}.~focus on reducing the computational time spent in fully-connected layer by imposing the circulant structure on the weight matrix for dimension reduction with little loss in performance \cite{iccv2015_ycheng_redundancycirculantprojection}. Ding \textit{et al}.~propose to use block-circulant structure in both fully-connected layers and convolutional layers in non-square-matrix situations to further reduce the storage waste. They also mathematically prove that fewer weights in circulant form do not harm the ability of a deep CNN without weight redundancy reduction \cite{ding2017c}.
}

\subsubsection{Knowledge Distillation}
\begin{figure}[tb!]
    \centering
    \includegraphics[height=0.4\textwidth]{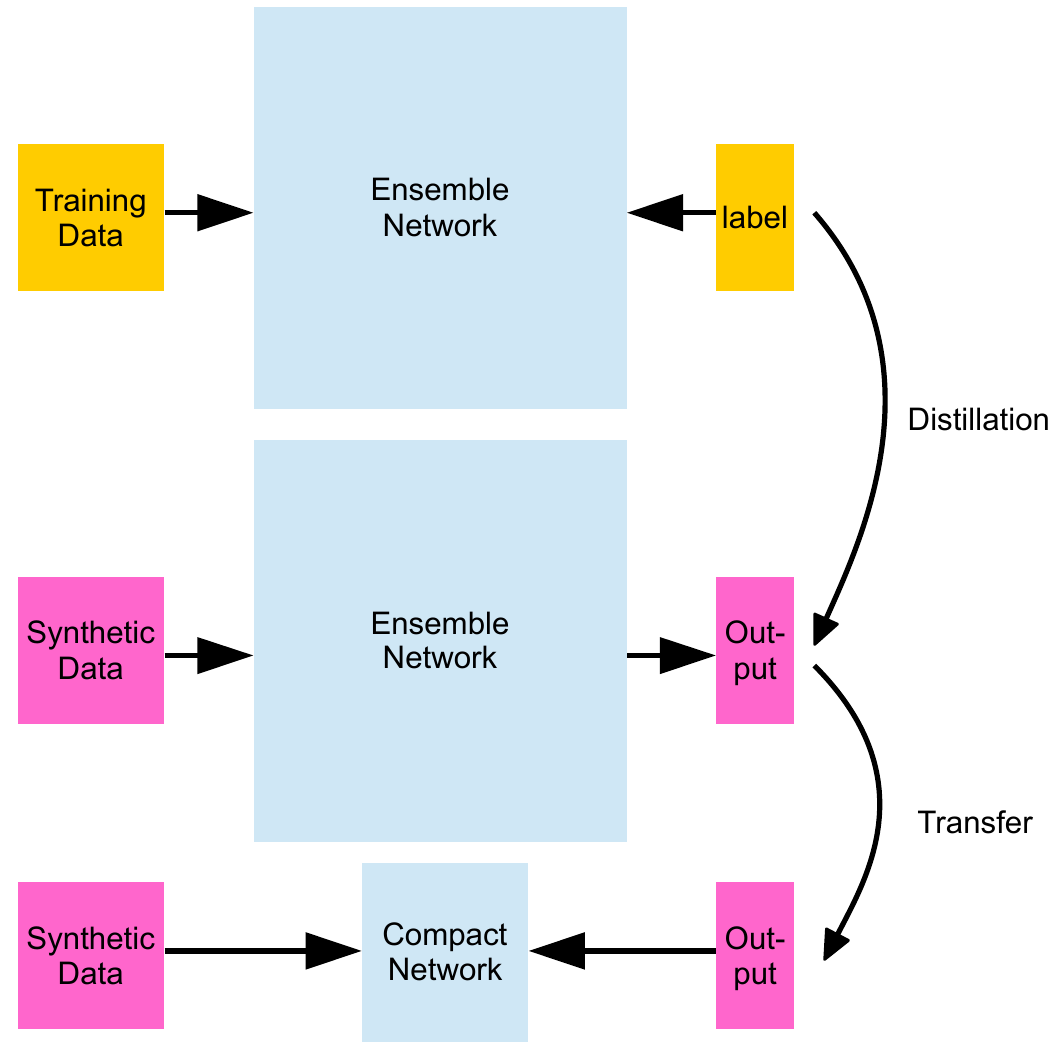}
    \caption{Illustration of knowledge distillation.}
    \label{fig:distillation}  
\end{figure}

Knowledge distillation is a concept that information obtained from a large complex ensemble neural networks can be utilized to form a compact neural network \cite{caruana2004ensemble}. The way that knowledge is transferred can be depicted in the following Figure~\ref{fig:distillation}. Information flow from one complex network to a simpler one by training the latter one with data labeled by the former network. By using synthetic data generated from a complex network to train a compact model, it is less likely to cause overfitting and can approximate the functions very well. More importantly, it provides a new perspective for model compression and complicated neural network acceleration. 

Synthetic data is very important in succeeding model compression. If it matches well with the true distribution from the functions of a complex model, it usually takes less data for training to mimic it with high-fidelity. Furthermore, the compact model has good generalization characteristics in some missions as it reduces overfitting. Bucilu \textit{et al}.~lay a foundation for mimicking a large machine learning model by experimenting three ways to generate pseudo data, which are random, naive bayes estimation, and MUNGE respectively \cite{acm2006_CBucila_modelCompres}. Some researches propose teacher-student format, which also adopts knowledge distillation concepts with different methods for synthesizing data. For example, Hinton \textit{et al}.~compress a deep teacher network into a student network using data combined from teacher network outcome and the true labeled data \cite{arxiv2015_ghinton_knowdistillation}. The student network can achieve very high accuracy on MNIST dataset with less run time of inference. Romero \textit{et al}.~mimic a wider and shallower teacher neural network with a thinner and deeper network called a student network by learning an intermediate representation that is predicted by the teacher network \cite{iclr2015_aromeo_fitnets}. The depth of the student network ensures its performance, while its thin characteristic reduces the computation complexity.

\subsection{Redundancy in Representations}
\label{subsec:represent}
Many weights in neural networks have very small values. For example, the first non-zero digit of many weights occurs in the eighth decimal place, which requires more precise way to record them. Most arithmetic operations in neural networks use 32-floating point representation in order to achieve a good accuracy. As a trade-off, that increases the computation workload and memory size for the neural networks. However, arithmetic operations in fixed-point instead of floating-point can achieve enough good performance for neural networks \cite{ijcnn1990_dhammerstrom_fixedarithmeticadequate}. A 16-bit fixed-point representation method is proposed by using stochastic rounding for training CIFAR-10 dataset  \cite{icml2015_sgupta_16bit}. A further compression of 10-bit dynamic fixed-point is also explored \cite{arxiv2014_mcourbariaux_10bits}. \add{Han \textit{et al}.~quantize pruned CNNs to 8-bit and achieve further storage reduction with no loss of accuracy \cite{iclr2016_shan_pruningquantizationhuffman}.}

For now, representation in \add{one bit} is the simplest form. In terms of binarization, there can be three forms, binary input, binary weights of the network, and binary operations. Courbariaux \textit{et al}.~propose a BinaryConnect method to use \add{1-bit} fixed-point weights to train a neural network \cite{NIPS2015_mcourbariaux_binaryweights}. Rastegari \textit{et al}.~come up with a XNOR-Nets with binary weights and binary input fed to convolutional layers \cite{eccv2016_mrastegari_xnor}. It results in $58\times$ speedup of convolutional operations. Kim \textit{et al}.~propose a Bitwise Neural Network, which takes everything as binary such as weights, bias terms, input, output, and basic logic operations instead of floating or fixed-point arithmetic operations \cite{icml2016_mkim_bitwisenn}. \add{Zhou \textit{et al}.~propose to train CNN using binary and stochastically quantized low bit-width gradients and achieve comparable performance as 32-bit counterparts \cite{zhou2018dorefa}. Hubara \textit{et al}.~propose training methods for quantized neural networks that use low precision including 1-bit weights and activations, and replace most arithmetic operations with bit-wise operations \cite{hubara2016quantized}. Kim \textit{et al}.~compress binary weight CNNs by decomposing kernels into sub-kernels with common parts. They reduce the operation of each image by $47.7\%$ \cite{kim2017kernel}. Ternary CNNs are proposed recently as a more expressive method comparing to binary CNNs, which seeks to achieve a balance between binary networks and full precision networks in terms of compression rate and accuracy \cite{li2016ternary,lin2016neural,alemdar2017ternary}.}

\add{Stochastic computing (SC) is a type of technique that simplifies numerical computations into bit-wise operations by representing continuous values with random bit streams. It provides many benefits for neural networks such as low computation footprint, error tolerance, simple implementation in circuits and better trade-off between time and accuracy \cite{brown2001stochastic}. Many works contribute to exploring potential space in optimization and in deep belief networks \cite{li2017structural,kim2016dynamic,ji2015hardware}. Recently it starts to gain attentions in CNN field and regarded as a promising technique for deep CNN implementation on ASIC (Section \ref{subsec:asic}) and on embedded portable devices as it can significantly reduce resource consumption with high accuracy.}

\add{SC is first adopted in deep CNN by Ren \textit{et al}.~with proposed method called SC-DCNN. They design both function blocks and feature extraction blocks that help SC efficiently implemented in deep CNN. It successfully achieves the lowest resource consumption of LeNet5 with optimized configurations among many state-of-the-art software and hardware platforms \cite{ren2017sc}. Li \textit{et al}.~further improve SC based DCNN by introducing normalization in the hardware implementation and dropout in DCNN software training. They design the stochastic normalization circuit by decoupling complex normalization into three units, namely, square and summation, activation and division. Their proposed method improves the SC-based DCNN with $3.26\%$ top-1 accuracy and $3.05\%$ top-5 accuracy \cite{li2017normalization}.}

Although errors may accumulate due to representation approximation, its hardware implementation can achieve a much faster speed and lead to less energy consumption.

\section{Algorithm Level}
\label{sec:opt}

In the training process, gradient-based method is widely used in multi-layer feedforward neural networks (FNN),
while some other models use analytically determined methods to minimize the cost function \cite{ncaa2014_li_determinedtrain}.
%As one type of FNN, the mainstream of training methods for CNN is gradient decent algorithm.
In the forward pass, the output of CNN is calculated, while in the backward pass, weights and bias are adjusted. By reducing the number of iterations to converge, training time can be decreased. Therefore, optimizing gradient decent algorithm is very important for improving the performance in training. For CNN, convolution computation reduces the amount of weights dramatically because it focuses on a local perception field. But repeated mathematic addition and multiplication increase the computation intensity. Therefore, in the forward process, convolution operation workloads are computationally intensive and become constrained for implementation. In the following, we discuss the algorithm optimization of the two directions of data flow, which are gradient-based backward training methods and convolution-based forward inference methods. \add{We summarize distributed gradient descent methods, hybrid variants of the gradient decent and the improvement in terms of self-adaptive learning rates, momentum factors, and partial gradients. We also give an overview on \add{im2col-based} algorithms, Winograd based algorithms, and FFT, \add{all of} which address the convolution cost problem in CNN.} 

\subsection{Gradient Decent Optimization}
\label{subsec:gd}
Gradient decent is one of the most popular algorithms for optimization.
It has been largely used in finding global minima for error functions during training neural networks, because it is simple and empirical to implement.
The core of \add{mathematical} model of gradient decent algorithm is the update rule $\bm{\theta} = \bm{\theta} - \eta \cdot \nabla_{\bm{\theta}}J(\bm{\theta})$,
where the parameters are updated in the opposite direction of the gradient of the error function $\nabla_{\bm{\theta}}J(\bm{\theta})$. 

\begin{figure*}[tb!]
    \centering
    \includegraphics[height=0.35\textwidth]{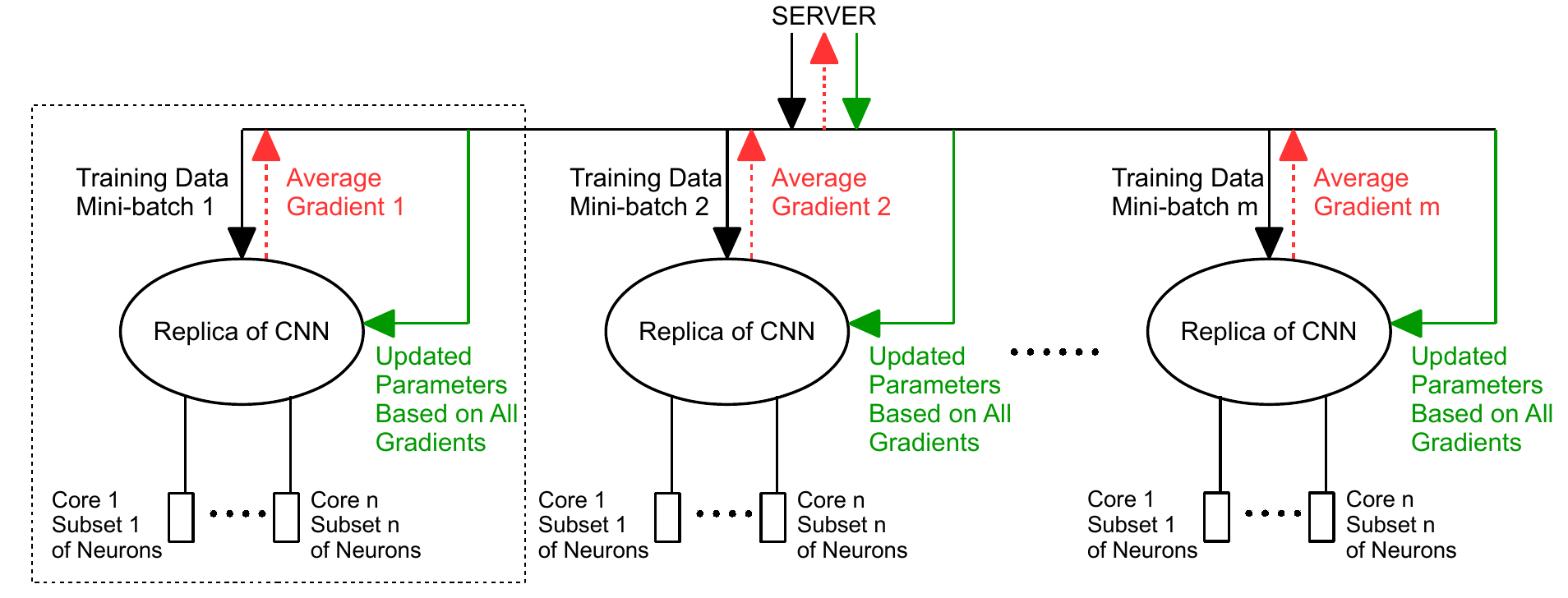}
    \caption{Illustration of CNN distributed system.}
    \label{fig:distributedCNN}  
\end{figure*}

\add{Distributed gradient decent methods have been proposed to alleviate hardware workload.
Take Google training CNN \cite{arxiv2015_ghinton_knowdistillation} as an example.
As illustrated in Figure~\ref{fig:distributedCNN}, there are two types of parallelism for the distribution.
(a) Replica of CNNs are trained through a server using averaging gradients and different batches of data.
Parameters are updated based on all the average gradients, which indicates that new parameters reflect the features from the whole data.
(b) For each replica of CNN, it distributes the computation into different cores with different subset of neurons.
Its implementation will be introduced in Section \ref{subsec:gpu}.}

Back-propagation algorithm is a form of gradient decent algorithm that is implemented in the neural networks. Some hybrid variants of back-propagation have been proposed in order to take advantage of the benefits from other algorithms. For example, combining it with cuckoo search algorithm can increase the searching speed for optimal solutions \cite{iccsa2013_nmohd_bpcuckoo}. The combination with ant colony algorithm can decrease the computational cost with increased stability of convergence \cite{procenn2006_yliu_antbp}. 
Pan \textit{et al}.~introduce three stages in the back-propagation with genetic algorithms and steepest decent methods combined together to achieve a fast and stable goal \cite{ncaa2014_pan_gawithbp}.
Ding \textit{et al}.~use genetic algorithms to optimize the weights of a back-propagation neural network by encoding and thresholding the connection weights \cite{ding2011optimizing}.

For deep CNN, as errors accumulate layer by layer, the gradient either decays rapidly to zero or increases out of bound. \add{Researchers focus on making changes in error functions, learning rates and incorporating momentum \cite{rumelhart1985learning} to reduce the derivative vanishing effects and to improve the speed of convergence in the 1990's, while in the recent 7 years, incorporating various factors with momentum factors, introducing self-adaptive learning rates, and using partial gradients are mainstreams to improve the gradient algorithms.}

For example, adapting learning rate to parameters with exponentially decaying average of squared gradients \add{leads} to a varying learning rate, which depends on each current and past parameter instead of being a constant \cite{jmlr2011_jduchi_adagrad,arxiv2012_mzeiler_adadelta,jmlr2015_dkingma_adam}. Hamid \textit{et al}.~incorporate the momentum factor and give control over it, which accelerates the convergence speed especially for oscillating situations in ravine \cite{ucma2011_nhamid_bpmomentum}. Nesterov \textit{et al}.~have proposed to use partial gradient to update each parameter rather than using the whole gradient \cite{siam2012_ynesterov_gdcoordinate,springer2014_prichtarik_gdblockcoordinate}. They randomly collect feature dimensions by sampling a block of coordinates and taking partial derivatives over this block, which can dramatically reduce the gradient computation complexity. As a result, it is much faster than the regular stochastic gradient decent method especially for high-dimensional dataset.

For deep neural networks, gradient descent with back-propagation is not guaranteed to find the global minimum of the error function, and is subject to weight vanishing or exploding. The former issue is due to the non-convexity of error functions in neural networks. Some works focus on non-gradient-based methods, such as ant bee colony algorithms and genetic algorithms. They are usually for simple dataset like Boolean dataset and simple neural network structures with one to two hidden layers. In practical, local minimum problem can be leveraged by a deep architecture \cite{lecun2015deep}. The second issue that weight vanishes or explodes when the amount of layers accumulates is still an open problem and has much potential to explore.

\subsection{Feed-forward Efficient Convolution}
\add{Three methods are summarized for the feed-forward efficient convolution including im2col-based algorithm, Winograd based method, and FFT based method, with the most commonly used one being introduced firstly. For the direct convolution in the CNN, convolution kernels slide over the two dimensions of the input and the output is obtained by dot product between the kernels and the input. While for the \add{im2col-based} algorithms, the input matrix is linearized into multiple lowered vectors, which can be later efficiently computed \cite{cublas,mlk,openblas}. Cho \textit{et al}.~further reduce the linearization memory-overhead and improve the computational efficiency by modifying both the lowered vectors and the vectorized kernels \cite{cho2017mec}. Winograd based methods are to incorporate Winograd’s minimal filtering algorithms to compute minimal convolution over small filters. Coppersmith Winograd algorithm is known as a fast matrix multiplication algorithm. Winograd based convolution reduces the multiplications by increasing the number of additions and it reduces the memory consumption on GPU \cite{cvpr2016_alavin_fftsmallfilter, park2016zero}.  \add{Winograd's minimal filtering algorithms can help reduce convolution computation at the expense of memory bandwidth. Xiao \textit{et al}.~utilize Winograd's minimal filtering theory combined with heterogeneous algorithms for a fusion architecture to mitigate memory bandwidth problem \cite{xiao2017exploring}.} }

\add{Based on the experiment that FFT can be applied in MLP to accelerate the first layer inference \cite{procavbpa1999_benyacoub_fftfirstlayer}, Mathieu \textit{et al}.~first apply FFT on weights of CNN and achieve good performance and speedup for large perceptive areas \cite{arxiv2014_mmathieu_trainingffts}. For using FFT in CNN, it is necessary to transform back and forth between time domain and frequency domain, which consumes a lot of resources and takes time. Therefore, it needs delicate balance between the benefits of computation in frequency domain and the drawbacks of transforming back and forth. Large perception areas have better performance, which results in limitation in the neural network with small convolution filters. In order to solve this problem, one of the solutions is to train weights directly in frequency domain \cite{neco2015_tbrosch_frequencydomain2d3d}. \add{Ko \textit{et al}.~train the CNNs entirely in the frequency domain with approximate frequency-domain nonlinear operations, sinc interpolation and Hermitian symmetry. By eliminating Fourier transforms at each layer, they achieve significantly training time reduction for CIFAR-10 recognition \cite{ko2017design}.}}

\section{Implementation Level}
\label{sec:acce}
Neural networks regain \add{their} vigor due to high performance hardware recently. CPU used to be the main stream for implementing machine learning algorithms about twenty years ago, \add{because} matrix multiplication and factorization techniques were not popular back then. Nowadays, GPU, FPGA, and ASIC are utilized for accelerating training and predicting process. Besides, much new device technology is proposed to meet requirement for very large models and large training datasets. In the following, hardware based accelerators are summarized in terms of GPU, FPGA, ASIC and frontier new device that is promising for accelerating deep convolutional neural networks.

\subsection{GPU}
\label{subsec:gpu}
In terms of GPU, clusters of GPUs can accelerate very large neural networks with over one billion parameters in a parallel way. The mainstream of GPU cluster neural networks usually work with distributed SGD algorithms as illustrated in Section \ref{subsec:gd}. 
Many researches further exploit the parallelism and make efforts on communication among different clusters. 
For example, Baidu Heterogeneous Computing Group \add{uses} two types of parallelism called model-data parallelism and data parallelism to extend CNN architectures to 36 servers, each with 4 NVIDIA Tesla K40m GPUs and 12GB memory. The strategies include butterfly synchronization and lazy update, which makes good use of overlapping in computation and communication \cite{wu2015deep}.
Coates \textit{et al}.~propose a clustering of GPU servers using Commodity Off-The-Shelf High Performance Computing (COTS HPC) technology and high-speed communication infrastructure for parallelism in distributed gradient decent algorithm, which reduces $98\%$ number of machines used for training \cite{coates2013deep}. \add{In terms of non-distributed SGD algorithms, Imani \textit{et al}.~propose a nearest content addressable memory block called NNCAM, which stores highly frequent patterns for reusing. It accelerates CNNs over general purpose GPU with $40\%$ speedup \cite{imani2017efficient}.}

\subsection{FPGA}
There are many parallelism levels in hardware acceleration, such as coarse-grain, medium-grain, fine-grain, and massive \cite{izeboudjen2014new}. FPGA outperforms in terms of its fine grain and coarse grain reconfiguration ability and its hierarchical storage structure and scheduling mechanism can be optimized flexibly. Flexible hierarchical memory systems can support complex data access mode of CNN. It is often used to improve the efficiency of on-chip memory and to reduce the energy consumption. 

Peemen \textit{et al}.~experiment on Virtex 6 FPGA board and show that the accelerator design can achieve $11\times$ speedup with very complicated address mapping of data access \cite{iccd2013_mpeemen_memorycentricaccelerator}. 
Zhang \textit{et al}.~take data reuse, parallel processing, and off-chip memory bandwidth into consideration in FPGA accelerator. The accelerator achieves $17.42\times$ faster speed than CPU in AlexNet CNN architecture \cite{procacm-sigda2015_czhang_optfpga}. 
Martínez \textit{et al}.~take advantage of the FPGA reconfiguration characteristics by unfolding the loop execution on different cascading stages. As the number of multipliers for convolution increases, the proposed method can achieve 12 GOPS at most \cite{martinez2013efficient}. 
A hardware acceleration method for CNN is proposed by combining fine grain in operator level parallelism and coarse grain parallelism. Compared with 4xIntel Xeon 2.3 GHz, 1.35 GHz C870, and a 200 MHz FPGA, the proposed design achieves a $4\times$ to $8\times$ speed boost \cite{procisca2010_chakradhar_dynamicconfigurab}. \add{Wang \textit{et al}.~propose an on-chip memory design called Memsqueezer that can be implemented on FPGA. They shrink the memory size by compressing data, weights, and intermediate data from the perspectives of hardware, which achieves $80\%$ energy reduction compared with conventional buffer designs \cite{wang2016re}. Zhang \textit{et al}.~design an FPGA accelerator engine called Caffeine that decreases underutilized memory bandwidth. It reorganizes the memory access according to their proposed matrix-multiplication representation applied to both convolutional layers and fully-connected layers. Caffeine’s implementation on Xilinx KU060 and Virtex 7690t FPGA achieves very high peak performance of 365 GOPS and 636 GOPS respectively \cite{zhang2016caffeine}. Rahman \textit{et al}.~present a 3D array architecture, which can benefit all layers in CNNs. With optimization of on-chip buffer sizes for FPGAs, it can outperform the state-of-the-art solutions by $22\%$ in terms of MAC \cite{rahman2016efficient}. Alwani \textit{et al}.~explore the design space of dataflow across multiple convolutional layers, where a fused layer accelerator is designed that reduces feature map data transfer from and to off-chip memory \cite{alwani2016fused}.}

\subsection{ASIC}
\label{subsec:asic}
For ASIC design, despite of using methods in structure level such as block-circulant projection in Section \ref{subsub:circulant} and SC in Section \ref{subsec:represent}, to improve the design in implementation level, memory can be expanded and locality can be increased to reduce data transporting within systems for deep neural network accelerating.
Tensor Processing Unit (TPU) is designed for low precision computation with high efficiency. It uses a large on-chip memory of 28MiB to execute the neural network applications, which can achieve at most $30\times$ faster speed than an Nvidia K80 GPU \cite{jouppi2017datacenter}. \add{TETRIS is an architecture using 3D memory proposed by Gao \textit{et al}. It saves more area for processing elements and leaves more space for accelerator design \cite{gao2017tetris}. }

Luo \textit{et al}.~create an architecture of 64-chip system that minimizes data moving between synapses and neurons by storing them closely. It reduces the burden on external memory bandwidth and achieves a speedup of $450\times$ over a GPU with $150\times$ energy reduction \cite{luo2017dadiannao}. \add{Wang \textit{et al}.~propose to group adjacent process engines (PEs) into dual-channel PEs called Chain-NN to mitigate huge amount of data movements. They simulate it under TSMC 28nm process and achieve a peak throughput of $806.4$ GOPS in AlexNet \cite{wang2017chain}.} Single instruction multiple data (SIMD) processors are used on a 32-bit CPU to design a system targeted for ASIC synthesis to perform real-time detection, recognition and segmentation of mega-pixel images. They optimize the operation in CNN with available parallelism in hardware. The ASIC implementations outperform the CPU conventional methods in terms of frames/s \cite{prociscas2010_cfarabet_hardwareparallelaccelr}. 

\add{Recently, some ASIC designs target at sparse networks with irregularity.
For example, Zhang \textit{et al}.~propose an accelerator called Cambricon-X that can reach $544$ GOP/s in $6.38mm^2$ \cite{zhang2016cambricon}.
It consists an Indexing Module, which can efficiently schedule processing elements that store irregular and compressed synapses.
Kwon \textit{et al}.~design a reconfigurable accelerator called MAERI to adapt various layer dataflow patterns.
They can efficiently utilize compute resources and provides $6.9\times$ speedup at $50\%$ sparsity \cite{kwon2018maeri}.
Network pruning could induce sparsity and irregularity as discussed in Section \ref{subsub:prune}. With such designs, better performance is expected to achieve when combined.}

\subsection{New Devices}
As new device technology and circuits arise, deep convolutional neural networks can be potentially accelerated by orders of magnitude.
In terms of new device, very large scale integration systems are explored to mimic complex biological neuron architectures. 

Some of them are in their theoretical demonstration state for training deep neural networks. For example, Gokmen and Vlasov from IBM research center propose a resistive processing unit (RPU) device, which can both store and compute parameters in this unit. It has extremely high processing speed with $30000\times$ higher than state-of-the-art microprocessors ($84000$ GigaOps/s/W) \cite{fnins2016_tgokmen_resistivedevice}. As neuromorphic engineering develops, more new devices emerge to handle high frequency and high volume information transformation through synapses. Some are  in theoretical state that have not been implemented on neural networks for classification and recognition, such as nano-scale phase change device \cite{jackson2013nanoscale} and ferroelectric memristors \cite{saighi2015plasticity}. 

\add{Resistive memories are treated as one of the promising solutions for deep neural network accelerations due to its nonvolatility, high storage density, and low power consumption \cite{seo2015chip}. Its architecture mimics neural networks, where weight storage and computation can be done simultaneously \cite{ncaa2016_zeng_memoristor,shim2016low}.} As CMOS memories become larger, its scale becomes limited. Therefore, besides the main stream CMOS based memory, nonvolatile memory becomes more popular in storing weights, such as resistive random access memory (RRAM) \cite{ni2016line,xu2014parallel,prezioso2015training,cheng2017time} and spin-transfer torque random access memory (STT-RAM) \cite{tvlsis2017_song_rambuffer}.

\add{Memristor crossbar array structures can deal with computational expensive matrix multiplication and have been explored in CNN hardware implementations.
For example, Hu \textit{et al}.~develop a Dot-Product Engine (DPE) utilizing memristor crossbar, which achieves $1000\times$ to $10,000\times$ speed-efficiency product compared with a digital ASIC \cite{hu2016dot}.
Xia \textit{et al}.~address energy consumption problem between crossbars and ADC/DAC and can save more than $95\%$ energy with similar accuracy of CNN \cite{xia2016switched}.
Ankit \textit{et al}.~propose a hierarchical reconfigurable architecture with memristive crossbar arrays called RESPARC,
which is $15\times$ more energy efficient and has $60\times$ more throughput for deep CNNs \cite{ankit2017resparc}. }

In general, for any CNN hardware implementation, there are a lot of potential solutions to be explored in design space.
It is not trivial to design a general hardware architecture that can be applied to every CNN, especially when limitations on computation resource and memory bandwidth are considered.

\section{Discussion}
\label{sec:discuss}
\add{Researches have different flavors over dataset, model, and implementation platforms.
Many datasets and models are treated as benchmarks based on previous researches. But different benchmarks and their combinations make it difficult to compare the method in one level with one in other levels.
In the following discussion, we constrain our comparison and analysis within each level.}

\subsection{Structure Level}
\begin{table*}[!htb]
    %{{{
	\centering
	\begin{tabular}{p{2.0cm}p{1.2cm}p{3.5cm}p{2.0cm}p{3.7cm}}
		%\hline
		\toprule
		~& Method & Target Layer & Pre-training & Performance \\
		\hline
		\hline
		\multirow{4}*{}{Layer decomposition} & \cite{jaderberg2014speeding} & Convolutional layers & required & $2.5\times$ speedup with no loss in accuracy \\
		\cline{2-5}
		~& \cite{nips2014_edentonylecun_linearstructureconvapprox} & Convolutional layers & required & $2\times$ speedup with $<1\%$ accuracy drop \\
		\cline{2-5}
		~& \cite{kim2015compression} & Whole network & required & $1.09\times$ reduction in weights \& $4.93\times$ speedup in VGG-16 \\
		\cline{2-5}
		~& \cite{ioannou2015training} & Convolutional layers & not required & $76\%$ reduction in weights in VGG-11 \\
		\hline
		\multirow{4}*{}{Pruning} & \cite{cvpr2015_bliu_sparsecnn} & \add{Whole network} & required & prune $90\%$ parameters \add{of the convolutional kernels}\\
		 \cline{2-5}
		~& \cite{nips2015_shan_prunconnections} & \add{Whole network} & required & prune $13\times$ parameters in VGG-16\\
		\cline{2-5}
		~& \cite{wen2016learning} & Whole network & not required & $5.1\times$ (CPU) \& $3.1\times$ (GPU) speedup in convolutional layers \\
		\cline{2-5}
		~& \cite{li2016pruning} & Whole network & required & $34\%$ inference \add{FLOP} reduction in VGG-16 \\
		\bottomrule
	\end{tabular}
	\caption{\add{Layer decomposition and pruning methods analysis}}
    %}}}
	\label{tab:train_inf}
\end{table*}

\add{We have summarized methods of layer decomposition and pruning in Table~\ref{tab:train_inf}.} Some of the layer decomposition and pruning methods focus on inference, because pre-trained CNNs are required before applying the corresponding methods. It is a limitation comparing to other acceleration methods. \add{For example, some large scale networks still need training for weeks or months before layer decomposition method implementation \cite{jaderberg2014speeding,nips2014_edentonylecun_linearstructureconvapprox}.} Pruning by sparsified weights and their connections require pre-training on the original full model and fine-tuning \cite{cvpr2015_bliu_sparsecnn,nips2015_shan_prunconnections}.

\add{Many layer decomposition and pruning methods are layer-wised and optimized for specific layer when they are first time proposed.} For example, Sainath \textit{et al}.~demonstrate significant reduction in parameters in the output softmax layer \cite{icassp2013_tsainath_lowrank}.
Mariet \textit{et al}.~successfully prune $25\%$ of the parameters with good performance in the fully connected layer \cite{iclr2016_ZMariet_mergeNeuronPruning}.
Denton \textit{et al}.~successfully reduce a large magnitude number of parameters in the convolutional layer \cite{nips2014_edentonylecun_linearstructureconvapprox}.
\add{As these methods focus on different types of layers, there is exploration space about how to combine them for further acceleration.}

After reducing redundancy in representation, the size of neural networks can reduce dramatically. However, specific hardware is required to achieve a speedup in training and testing, since currently most GPUs are improved to suit for floating-point performance. For example, using BinaryConnect method \cite{arxiv2016_mcourbariaux_binaryweightsactivations} to train a Torch7 frame \add{ConvNet} on GPU takes more time. The time complexity can be reduced theoretically by $60\%$ if using dedicated hardware. 

\begin{table*}[!htb]
    %{{{
	\centering
	\begin{tabular}{p{1.5cm}p{2.5cm}p{2.0cm}p{2.3cm}p{4.2cm}}
		\toprule
		Bit-width & Method & Model & Error rate increase & \add{Note} \\
		\hline
		\hline
        \multirow{3}*{}{Binary} & BinaryConnect \cite{NIPS2015_mcourbariaux_binaryweights} & Self-designed (e.g.~\add{6C-3S-2F-L2SVM}) & $\sim 2\%$ error rate drop & Binary weights during training and testing \\
        \cline{2-5}
        ~& Binarynet \cite{arxiv2016_mcourbariaux_binaryweightsactivations} & Self-designed & \add{$10.15\%$ absolute error rate} & Binary weights \& activations during forward pass\\
        \hline
        \multirow{3}*{}{Ternary} & TWN \cite{li2016ternary} & VGG-7 & $<1\%$ & Ternary weights in forward \& backward pass \\
        \cline{2-5}
        ~& Ternary Connect \cite{lin2016neural} & 6C-1\add{F}-1\add{softmax} & $\sim 3\%$ error rate drop & Ternary weights during training \\
        \cline{2-5}
        ~& TNN \cite{alemdar2017ternary} & VGG-variant & $12.11\%$ absolute error rate & Teacher-student approach based ternary \add{input} \& activations during training \\
        \hline
        \multirow{2}*{}{Others} & 12/14/16-bit \cite{icml2015_sgupta_16bit} & 3C-3S-1softmax & $< 5\%$ & Fixed-point number representation with stochastic rounding \\
        \cline{2-5} 
        ~& 10-bit \cite{arxiv2014_mcourbariaux_10bits} & Maxout networks & $<5\%$ & \add{Dynamic 10-bit fixed point precision during training} \\
		\bottomrule
	\end{tabular}
	\captionsetup{justification=centering}
	\caption[caption]{\add{Representation Reduction Methods (CIFAR10) \hspace{\textwidth} C: convolutional layer, S: subsampling layer, F: fully-connected layer}}
    %}}}
	\label{tab:cifar10}
\end{table*}

\begin{table*}[!htb]
    %{{{
	\centering
	\begin{tabular}{p{1.0cm}p{2.0cm}p{2.0cm}p{1.5cm}p{4.5cm}}
		\toprule
		Bit-width & Method & Model & Error rate increase & \add{Note} \\
		\hline
		\hline
		\multirow{4}*{}{Binary} & QNN \cite{hubara2016quantized} & GoogLeNet, AlexNet & $>10\%$ & Binary weights \& activations during training and testing \\
		\cline{2-5}
		~ & XNOR-net \cite{eccv2016_mrastegari_xnor} & AlexNet, ResNet, GoogLeNet-variation & $>10\%$ & Binary weights \& input to convolutional layers \\
		\cline{2-5}
		~ & BWN \cite{eccv2016_mrastegari_xnor} & AlexNet, ResNet, GoogLeNet-variation & $<10\%$ & Binary weights in forward \& backward pass\\
		\cline{2-5}
		~ & DoReFa-Net \cite{zhou2018dorefa} & AlexNet & around $10\%$ & Binary weights \& 2-bit activations \& 6-bit gradients \\
		\hline
		Ternary & TWN \cite{li2016ternary} & ResNet & $<5\%$ & Ternary weights in forward \& backward pass \\
		\bottomrule
	\end{tabular}
	\caption{\add{Representation Reduction Methods (ImageNet).}}
    %}}}
	\label{tab:imagenet}
\end{table*}

\add{We have summarized methods of reducing redundancy in representation in Table~\ref{tab:cifar10} and Table~\ref{tab:imagenet}. The Note column in these two tables provide important information that needs to be distinguished among different methods. Low-bit representation methods are targeted for both large and small models. Various bit-width representation results in different performance dependent on different models and datasets. Many experiments are conducted based on small datasets with low resolution (e.g.~$32\times 32$) such as CIFAR10, and usually achieve less than $5\%$ error rate increase. For image classification at large scale (e.g.~ImageNet), low-bit representation method is difficult to achieve the same performance as that for small dataset.}

\subsection{Algorithm Level}
\subsubsection{Information for Updating}
According to the cost function of gradient descent algorithm, there are two categories, which are first-order derivatives gradient descent and second-order derivatives gradient decent. Compared with first-order derivatives based gradient decent algorithm, the second-order one is faster to converge. But it is more complex to utilize the second order information, which makes it prohibitive in practice for deep large neural networks. Therefore, more emphasis has been put on how to approximate the Hessian matrices, which consists of the second-order derivatives for simplicity \cite{liew2016optimized}. 

\subsubsection{Data for Training}
According to the update amount of data, there are three variants of the algorithms, which are batch, mini-batch, and online. The batch method uses whole dataset to update the gradient in one iteration. The mini-batch uses randomly picked small amount of data to update the gradient while the online method uses new incoming subset of data once to update the gradient and stops at any time. 

For batch gradient decent, it is guaranteed to converge to the global minimum for convex surfaces. But it can be very slow and requires very large memory storage. The mini-batch can avoid redundant gradient computation using shuffled examples. As a result, it usually shoots the minimum faster than the batch gradient. With delicate picked learning rate, its fluctuation performance decreases. The online method can be used for designing a light-weight algorithm in terms of memory and speed for handling a stream of data. Since the data is updated frequently, it can be used to predict the most recent state of the trend. But as data is discarded after gradient update, online method is considered to be more difficult and unreliable \cite{cho2014co}. 

\subsubsection{Asynchronous Updating}
Asynchronous training algorithms help improve the efficiency on large-scale clusters of machines for distributed training and inference. Proposed asynchronous stochastic gradient decent algorithms such as Downpour SGD and AASGD help improve neural network performance in classification problems with a large amount of high dimensional data. But more attention is needed on communication among different workers in the clusters, since suboptimal communication can result in parameters diverging \cite{dean2012large,meng2016asynchronous}. 

\begin{table*}[!htb]
    %{{{
	\centering
	\begin{tabular}{p{1.0cm}p{1.5cm}p{2.3cm}p{1.5cm}p{2.5cm}p{3.6cm}}
		\toprule
		Method & Platform & Feature & Additional Memory & Complexity of Fourier Transform \& Inverse & Complexity of Add \& Mul in Frequency Domain \& Extra Complexity \\
		\hline
		\hline
		Mathieu \cite{arxiv2014_mmathieu_trainingffts} & GeForce GTX Titan GPU & Perform convolutions as products in frequency domain & Yes & \multirow{2}*{}{$(2C\cdot n^2\log n)(S\cdot f + f' \cdot f + S \cdot f')$} & \multirow{2}*{}{$4S \cdot f' \cdot f \cdot n^2$} \\
		\cline{1-4}
		Rippel \cite{nips2015_orippel_dftinpoolingcnn} & Xeon Phi coprocessor & Pooling in frequency domain & Yes &  &  \\
		\hline
		Ko \cite{ko2017design} & ASIC & Train the network entirely in frequency domain & No & $2Cn^2\log n(S \cdot f)$ & $(3/2 \cdot (n^2-\alpha)+\alpha) \cdot S \cdot f' \cdot f + 3/2 \cdot n^2 \cdot k^2 \cdot f' \cdot f $ \\
		\hline
		\multicolumn{6}{c}{S: mini-batch size, f: input feature map depth, f': output feature map depth, \add{n: feature map dimension}}\\
		\multicolumn{6}{c}{k: kernel dimension, C: hidden constant in the $\mathcal{O}$ notation, $\alpha$: 1 for odd and 4 for even n}\\ 
		\bottomrule
	\end{tabular}
	\caption{\add{FFT based method analysis.}}
    %}}}
	\label{tab:fft}
\end{table*}

\subsubsection{From Frequency Perspective}

\begin{table*}[!htb]
    %{{{
	\centering
	\begin{tabular}{p{3.5cm}p{1.5cm}p{3.0cm}p{1.5cm}p{2.8cm}}
		\toprule
		Method & Platform & Memory & Frequency & Performance \\
		\hline
		\hline
		Minwa \cite{wu2015deep} & GPUs & $6.9$ TB host memory \& $1.7$ TB device memory & - & $0.6$ PFlops single precision at peak \\
		\hline
		\add{Roofline-model-based accelerator} \cite{procacm-sigda2015_czhang_optfpga} & VC707 FPGA  & - & $100$MHz & $61.62$ GFLOPS \\
		\hline
		Caffeine \cite{zhang2016caffeine} & Xilinx KU060 FPGA & - & $200$MHz & $365$ GOPS \\
		\hline
		\add{ICAN accelerator} \cite{rahman2016efficient} & Virtex-7 FPGA & Memory bandwidth $6.2$ GB/s & $160$MHz & $147.82$ GOPS \\
		\hline
		Dadiannao \cite{luo2017dadiannao} & ASIC & $36$MB node eDRAM & $606$ MHz & $2.09$ TeraOps/s of a node at peak \\
		\hline
		Chain-NN \cite{wang2017chain} & ASIC & $352$KB on-chip memory & $700$MHz & $806.4$ GOPS at peak \\
		\hline
		Cambricon-X \cite{zhang2016cambricon} & ASIC & $56$KB on-chip SRAM & $1$ GHz & $544$ GOPS \\
		\bottomrule
	\end{tabular}
	\caption{\add{Performance comparison among GPU, FPGA, and ASIC.}}
    %}}}
	\label{tab:gpu_fpga_asic}
\end{table*}

\add{Table~\ref{tab:fft} summarizes methods of FFT based CNNs.} The concept of implementing CNN in frequency is to replace convolution operation in time domain with multiplication in frequency domain. It takes time to transform back and forth. As a result, it performs well on large feature maps. Development is made to suit for small feature maps such as training network directly in frequency domain. Compared with other algorithms, FFT method requires additional memory for padding the filters to the same size of the input and storing frequency domain intermediate results. This leads to a trade-off for hardware implementation. On one hand, it can take use of power in GPU parallelism to speedup convolution computation dramatically. On the other hand, more delicate GPU memory allocation is required due to limit memory. 

\begin{figure}[tb!]
\begin{center}
\includegraphics[width=0.4\textwidth]{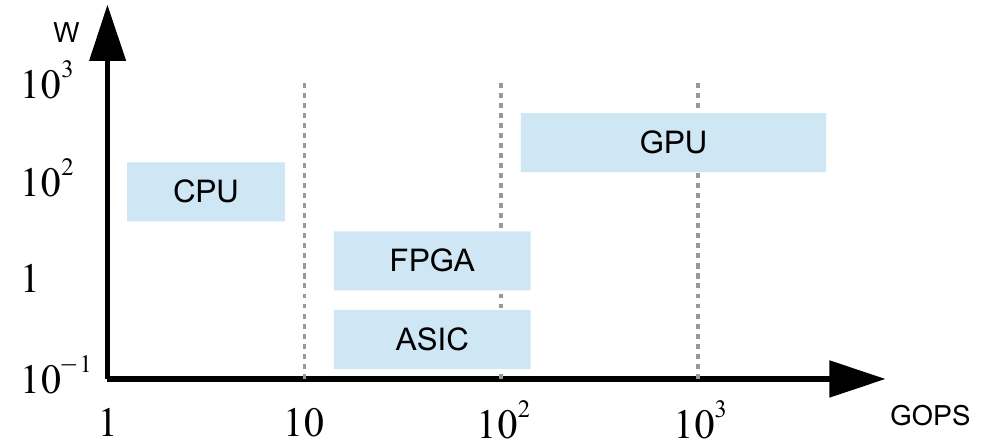}
\end{center}
\caption{Power and throughput among CPU, GPU, FPGA, and ASIC.}
\label{fig:powervsthroughput}      
\end{figure}

\subsection{Implementation Level}
As CNN becomes more and more complex, general purpose processors cannot exploit the inherent parallelism for matrix or tensor operations and therefore becomes bottleneck when performing large deep convolutional neural networks. Various designs for accelerating network training and inference have been proposed based on GPU, ASIC, and FPGA. \add{Table \ref{tab:gpu_fpga_asic} gives a performance comparison of different methods implementing on GPU, FPGA, and ASIC.} Figure~\ref{fig:powervsthroughput} shows the comparison among CPU, GPU, FPGA, and ASIC in terms of power and throughput \cite{du2017accelerator,han2016eie,qiu2016going,moons2017energy,chen2017eyeriss}.

GPU supports several teraFLOPS throughput and large memory access, but consumes a lot of energy. In terms of economy, GPU costs to set up for large deep convolutional neural networks. 

Comparing to GPU, ASIC is specialized hardware and can be delicately designed to maximize its benefits such as power-efficiency and large throughput in CNN implementation. However, once CNN algorithms are implemented on ASIC, it is difficult to change the hardware design. On the other hand FPGA is easy to be programmed and reconfigured. It is more convenient for prototyping.

Compared with GPU, FPGA throughput is tens of gigaFLOPS and it has limited memory access.
\add{In addition}, it does not support floating-point natively. But it is more power-efficient. Due to its limited memory access, many proposed methods are focused on accelerating inference time of neural network since inference process requires less memory access comparing to training process. Others are emphasized on external memory optimization for large neural network acceleration. Different models need different hardware optimization and even for the same model, different designs result in quite various acceleration performance \cite{iclr2016_shan_pruningquantizationhuffman}. In terms of economy, FPGA is reconfigurable and is easier to evolve hardware, frameworks and software. Especially for various models of neural networks, its flexibility shortens design cycle and costs less.

\section{Conclusion}
\label{sec:conclu}

In this paper, we \add{have summarized} recent advances in CNN acceleration methods \add{from all} structure level, algorithm level, and implementation level.
In structure level, CNN is compressed without losing significant accuracy since there is redundancy in most of the CNN architectures.
For training algorithms, besides convergence speed, convolution calculation is also an important factor for CNN.
FFT method introduces a frequency perspective for training neural networks.
In implementation level, characteristics for different hardware such as FPGA and GPU are explored combined with CNN features.
CNN performs better in computer vision field as its structure goes deeper and the amount of data becomes larger, which makes it time consuming and computationally expensive.
It is imperative and necessary to accelerate CNN for its further implementation in life.
For now, there is no generalized evaluation system to test the acceleration performance for comparison among different methods in different levels.
Researches use case by case dataset benchmark and different criterion in each level. Therefore, it is challenging in acceleration performance evaluation as well.

\section*{Acknowledgement}
This research work was partly supported by the Natural Science Foundation of China and Jiangsu (Project No.~6171101145, 61671148, BK20161147), the Research and Innovation Program for Graduate Students in Universities of Jiangsu Province (Grant No.~SJLX16\_0081, SJCX17\_0048), and the Research Grants Council of Hong Kong SAR (Project No.~CUHK24209017).

%\section*{References}
\bibliographystyle{elsarticle-num} 
\bibliography{./ref/Top,./ref/cnn} 

\begin{thebibliography}{100}
\expandafter\ifx\csname url\endcsname\relax
  \def\url#1{\texttt{#1}}\fi
\expandafter\ifx\csname urlprefix\endcsname\relax\def\urlprefix{URL }\fi
\expandafter\ifx\csname href\endcsname\relax
  \def\href#1#2{#2} \def\path#1{#1}\fi

\bibitem{yu2017convolutional}
S.~Yu, S.~Jia, C.~Xu, Convolutional neural networks for hyperspectral image
  classification, Neurocomputing 219 (2017) 88--98.

\bibitem{farabet2013learning}
C.~Farabet, C.~Couprie, L.~Najman, Y.~LeCun, Learning hierarchical features for
  scene labeling, IEEE Transactions on Pattern Analysis and Machine
  Intelligence 35~(8) (2013) 1915--1929.

\bibitem{wang2016parallel}
K.~Wang, C.~Gou, N.~Zheng, J.~M. Rehg, F.~Y. Wang, Parallel vision for
  perception and understanding of complex scenes: methods, framework, and
  perspectives, Artificial Intelligence Review (2016) 1--31.

\bibitem{qin2016empirical}
P.~Qin, W.~Xu, J.~Guo, An empirical convolutional neural network approach for
  semantic relation classification, Neurocomputing 190 (2016) 1--9.

\bibitem{yu2015machine}
B.~Yu, D.~Z. Pan, T.~Matsunawa, X.~Zeng, Machine learning and pattern matching
  in physical design, in: IEEE/ACM Asia and South Pacific Design Automation
  Conference (ASPDAC), 2015, pp. 286--293.

\bibitem{iclr2017_theis_lossy}
L.~Theis, W.~Shi, A.~Cunningham, F.~Husz{\'a}r, Lossy image compression with
  compressive autoencoders, in: International Conference on Learning
  Representations (ICLR), 2017.

\bibitem{zhang2016image}
M.~Zhang, T.~Chen, X.~Shi, P.~Cao, Image arbitrary-ratio down-and up-sampling
  scheme exploiting dct low frequency components and sparsity in high frequency
  components, IEICE Transactions on Information and Systems 99~(2) (2016)
  475--487.

\bibitem{chen2016image}
T.~Chen, M.~Zhang, J.~Wu, C.~Yuen, Y.~Tong, Image encryption and compression
  based on kronecker compressed sensing and elementary cellular automata
  scrambling, Optics \& Laser Technology 84 (2016) 118--133.

\bibitem{jothi2017survey}
J.~A.~A. Jothi, V.~M.~A. Rajam, A survey on automated cancer diagnosis from
  histopathology images, Artificial Intelligence Review 48~(1) (2017) 31--81.

\bibitem{jouppi2017datacenter}
N.~P. Jouppi, C.~Young, N.~Patil, D.~Patterson, G.~Agrawal, R.~Bajwa, S.~Bates,
  S.~Bhatia, N.~Boden, A.~Borchers, et~al., In-datacenter performance analysis
  of a tensor processing unit, in: IEEE/ACM International Symposium on Computer
  Architecture (ISCA), 2017, pp. 1--12.

\bibitem{nvdla}
{NVDLA}, \url{http://nvdla.org}.

\bibitem{nnp}
\url{https://www.intelnervana.com}.

\bibitem{guo2016deep}
Y.~Guo, Y.~Liu, A.~Oerlemans, S.~Lao, S.~Wu, M.~S. Lew, Deep learning for
  visual understanding: A review, Neurocomputing 187 (2016) 27--48.

\bibitem{ghayoumi2017quick}
M.~Ghayoumi, A quick review of deep learning in facial expression, Journal of
  Communication and Computer 14 (2017) 34--38.

\bibitem{carrio2017review}
A.~Carrio, C.~Sampedro, A.~Rodriguez-Ramos, P.~Campoy, A review of deep
  learning methods and applications for unmanned aerial vehicles, Journal of
  Sensors 2017.

\bibitem{zhang2015deep}
J.~Zhang, C.~Zong, Deep neural networks in machine translation: An overview,
  IEEE Intelligent Systems 30~(5) (2015) 16--25.

\bibitem{singh2017machine}
S.~P. Singh, A.~Kumar, H.~Darbari, L.~Singh, A.~Rastogi, S.~Jain, Machine
  translation using deep learning: An overview, in: International Conference on
  Computer, Communications and Electronics (Comptelix), 2017, pp. 162--167.

\bibitem{ling2015deep}
Z.~H. Ling, S.~Y. Kang, H.~Zen, A.~Senior, M.~Schuster, X.-J. Qian, H.~M. Meng,
  L.~Deng, Deep learning for acoustic modeling in parametric speech generation:
  A systematic review of existing techniques and future trends, IEEE Signal
  Processing Magazine 32~(3) (2015) 35--52.

\bibitem{schmidhuber2015deep}
J.~Schmidhuber, Deep learning in neural networks: An overview, Neural Networks
  61 (2015) 85--117.

\bibitem{lecun2015deep}
Y.~LeCun, Y.~Bengio, G.~Hinton, Deep learning, Nature 521~(7553) (2015)
  436--444.

\bibitem{du2016overview}
X.~Du, Y.~Cai, S.~Wang, L.~Zhang, Overview of deep learning, in: Youth Academic
  Annual Conference of Chinese Association of Automation (YAC), IEEE, 2016, pp.
  159--164.

\bibitem{pieee1998_LeCun_lenet}
Y.~Lecun, L.~Bottou, Y.~Bengio, P.~Haffner, {Gradient-based learning applied to
  document recognition}, Proceedings of the IEEE 86~(11) (1998) 2278--2324.

\bibitem{nips2012_krizhevsky_alexnet}
A.~Krizhevsky, I.~Sutskever, G.~E. Hinton, Imagenet classification with deep
  convolutional neural networks, in: Conference on Neural Information
  Processing Systems (NIPS), 2012, pp. 1097--1105.

\bibitem{lin2013network}
M.~Lin, Q.~Chen, S.~Yan, Network in network, arXiv preprint arXiv:1312.4400.

\bibitem{iclr2015_simonyan_vgg}
K.~Simonyan, A.~Zisserman, Very deep convolutional networks for large-scale
  image recognition, in: International Conference on Learning Representations
  (ICLR), 2015.

\bibitem{cvpr2016_he_resnet}
K.~He, X.~Zhang, S.~Ren, J.~Sun, Deep residual learning for image recognition,
  in: IEEE Conference on Computer Vision and Pattern Recognition (CVPR), 2016,
  pp. 770--778.

\bibitem{cvpr2015_szegedy_googlenet}
C.~Szegedy, W.~Liu, Y.~Jia, P.~Sermanet, S.~Reed, D.~Anguelov, D.~Erhan,
  V.~Vanhoucke, A.~Rabinovich, Going deeper with convolutions, in: IEEE
  Conference on Computer Vision and Pattern Recognition (CVPR), 2015, pp. 1--9.

\bibitem{chollet2017xception}
F.~Chollet, Xception: Deep learning with depthwise separable convolutions, in:
  IEEE Conference on Computer Vision and Pattern Recognition (CVPR), 2017.

\bibitem{aloysius2017review}
N.~Aloysius, M.~Geetha, A review on deep convolutional neural networks, in:
  Communication and Signal Processing (ICCSP), 2017 International Conference
  on, IEEE, 2017, pp. 0588--0592.

\bibitem{al2017review}
A.~A.~M. Al-Saffar, H.~Tao, M.~A. Talab, Review of deep convolution neural
  network in image classification, in: Radar, Antenna, Microwave, Electronics,
  and Telecommunications (ICRAMET), 2017 International Conference on, IEEE,
  2017, pp. 26--31.

\bibitem{rawat2017deep}
W.~Rawat, Z.~Wang, Deep convolutional neural networks for image classification:
  A comprehensive review, Neural computation 29~(9) (2017) 2352--2449.

\bibitem{bishop2006pattern}
C.~M. Bishop, {Pattern Recognition and Machine Learning}, Springer, New York,
  2006.

\bibitem{nair2010rectified}
V.~Nair, G.~E. Hinton, Rectified linear units improve restricted boltzmann
  machines, in: International Conference on Machine Learning (ICML), 2010, pp.
  807--814.

\bibitem{maas2013rectifier}
A.~L. Maas, A.~Y. Hannun, A.~Y. Ng, Rectifier nonlinearities improve neural
  network acoustic models, in: International Conference on Machine Learning
  (ICML), 2013.

\bibitem{he2015delving}
K.~He, X.~Zhang, S.~Ren, J.~Sun, Delving deep into rectifiers: Surpassing
  human-level performance on imagenet classification, in: IEEE International
  Conference on Computer Vision (ICCV), 2015, pp. 1026--1034.

\bibitem{xu2015empirical}
B.~Xu, N.~Wang, T.~Chen, M.~Li, Empirical evaluation of rectified activations
  in convolutional network, in: International Conference on Machine Learning
  Workshop, 2015.

\bibitem{clevert2015fast}
D.-A. Clevert, T.~Unterthiner, S.~Hochreiter, Fast and accurate deep network
  learning by exponential linear units (elus), in: International Conference on
  Learning Representations (ICLR), 2016.

\bibitem{sermanet2012convolutional}
P.~Sermanet, S.~Chintala, Y.~LeCun, Convolutional neural networks applied to
  house numbers digit classification, in: IEEE International Conference on
  Pattern Recognition (ICPR), 2012, pp. 3288--3291.

\bibitem{zeiler2013stochastic}
M.~D. Zeiler, R.~Fergus, Stochastic pooling for regularization of deep
  convolutional neural networks, in: International Conference on Learning
  Representations (ICLR), 2013.

\bibitem{yu2014mixed}
D.~Yu, H.~Wang, P.~Chen, Z.~Wei, Mixed pooling for convolutional neural
  networks, in: International Conference on Rough Sets and Knowledge
  Technology, Springer, 2014, pp. 364--375.

\bibitem{he2015spatial}
K.~He, X.~Zhang, S.~Ren, J.~Sun, Spatial pyramid pooling in deep convolutional
  networks for visual recognition, IEEE Transactions on Pattern Analysis and
  Machine Intelligence 37~(9) (2015) 1904--1916.

\bibitem{rippel2015spectral}
O.~Rippel, J.~Snoek, R.~P. Adams, Spectral representations for convolutional
  neural networks, in: Conference on Neural Information Processing Systems
  (NIPS), 2015, pp. 2449--2457.

\bibitem{gong2014multi}
Y.~Gong, L.~Wang, R.~Guo, S.~Lazebnik, Multi-scale orderless pooling of deep
  convolutional activation features, in: European Conference on Computer Vision
  (ECCV), Springer, 2014, pp. 392--407.

\bibitem{liu2017survey}
W.~Liu, Z.~Wang, X.~Liu, N.~Zeng, Y.~Liu, F.~E. Alsaadi, A survey of deep
  neural network architectures and their applications, Neurocomputing 234
  (2017) 11--26.

\bibitem{nips2013_mdenil_redundancyinnn}
M.~Denil, B.~Shakibi, L.~Dinh, N.~de~Freitas, et~al., Predicting parameters in
  deep learning, in: Conference on Neural Information Processing Systems
  (NIPS), 2013, pp. 2148--2156.

\bibitem{icassp2013_tsainath_lowrank}
T.~N. Sainath, B.~Kingsbury, V.~Sindhwani, E.~Arisoy, B.~Ramabhadran, Low-rank
  matrix factorization for deep neural network training with high-dimensional
  output targets, in: IEEE International Conference on Acoustics, Speech and
  Signal Processing (ICASSP), 2013, pp. 6655--6659.

\bibitem{jaderberg2014speeding}
M.~Jaderberg, A.~Vedaldi, A.~Zisserman, Speeding up convolutional neural
  networks with low rank expansions, in: British Machine Vision Conference
  (BMVC), 2014.

\bibitem{nips2014_edentonylecun_linearstructureconvapprox}
E.~L. Denton, W.~Zaremba, J.~Bruna, Y.~LeCun, R.~Fergus, Exploiting linear
  structure within convolutional networks for efficient evaluation, in:
  Conference on Neural Information Processing Systems (NIPS), 2014, pp.
  1269--1277.

\bibitem{iclr2015_vlebedev_cpdecomposition}
V.~Lebedev, Y.~Ganin, M.~Rakhuba, I.~Oseledets, V.~Lempitsky, Speeding-up
  convolutional neural networks using fine-tuned cp-decomposition, in:
  International Conference on Learning Representations (ICLR), 2014.

\bibitem{tai2015convolutional}
C.~Tai, T.~Xiao, Y.~Zhang, X.~Wang, et~al., Convolutional neural networks with
  low-rank regularization, arXiv preprint arXiv:1511.06067.

\bibitem{wang2016accelerating}
P.~Wang, J.~Cheng, Accelerating convolutional neural networks for mobile
  applications, in: ACM International Multimedia Conference (MM), 2016, pp.
  541--545.

\bibitem{kim2015compression}
Y.-D. Kim, E.~Park, S.~Yoo, T.~Choi, L.~Yang, D.~Shin, Compression of deep
  convolutional neural networks for fast and low power mobile applications, in:
  International Conference on Learning Representations, 2016.

\bibitem{ding2017compact}
H.~Ding, K.~Chen, Y.~Yuan, M.~Cai, L.~Sun, S.~Liang, Q.~Huo, A compact
  {CNN-DBLSTM} based character model for offline handwriting recognition with
  tucker decomposition, in: IAPR International Conference on Document Analysis
  and Recognition (ICDAR), Vol.~1, IEEE, 2017, pp. 507--512.

\bibitem{le2013fastfood}
Q.~Le, T.~Sarl{\'o}s, A.~Smola, Fastfood-approximating kernel expansions in
  loglinear time, in: International Conference on Machine Learning (ICML),
  Vol.~85, 2013.

\bibitem{iclr2016_mmoczulski_acdclinearlayer}
M.~Moczulski, M.~Denil, J.~Appleyard, N.~de~Freitas, {ACDC}: A structured
  efficient linear layer, in: International Conference on Learning
  Representations (ICLR), 2016.

\bibitem{ioannou2015training}
Y.~Ioannou, D.~Robertson, J.~Shotton, R.~Cipolla, A.~Criminisi, Training {CNNs}
  with low-rank filters for efficient image classification, in: International
  Conference on Learning Representations (ICLR), 2016.

\bibitem{wen2017coordinating}
W.~Wen, C.~Xu, C.~Wu, Y.~Wang, Y.~Chen, H.~Li, Coordinating filters for faster
  deep neural networks, in: IEEE International Conference on Computer Vision
  (ICCV), 2017.

\bibitem{tpami2015_asironi_separablefilters}
R.~Rigamonti, A.~Sironi, V.~Lepetit, P.~Fua, Learning separable filters, IEEE
  Transactions on Pattern Analysis and Machine Intelligence.

\bibitem{tpami2016_xzhang_gsvd}
X.~Zhang, J.~Zou, K.~He, J.~Sun, Accelerating very deep convolutional networks
  for classification and detection, IEEE Transactions on Pattern Analysis and
  Machine Intelligence 38~(10) (2016) 1943--1955.

\bibitem{iclr2016_shan_pruningquantizationhuffman}
S.~Han, H.~Mao, W.~J. Dally, Deep compression: Compressing deep neural networks
  with pruning, trained quantization and huffman coding, in: International
  Conference on Learning Representations (ICLR), 2016.

\bibitem{zhou2016less}
H.~Zhou, J.~M. Alvarez, F.~Porikli, Less is more: Towards compact {CNNs}, in:
  European Conference on Computer Vision (ECCV), Springer, 2016, pp. 662--677.

\bibitem{iclr2016_ZMariet_mergeNeuronPruning}
Z.~Mariet, S.~Sra, Diversity networks, in: International Conference on Learning
  Representations (ICLR), 2016.

\bibitem{access2015_apolyak_facenetspruning}
A.~Polyak, L.~Wolf, Channel-level acceleration of deep face representations,
  IEEE Access 3 (2015) 2163--2175.

\bibitem{wen2016learning}
W.~Wen, C.~Wu, Y.~Wang, Y.~Chen, H.~Li, Learning structured sparsity in deep
  neural networks, in: Conference on Neural Information Processing Systems
  (NIPS), 2016, pp. 2074--2082.

\bibitem{he2017channel}
Y.~He, X.~Zhang, J.~Sun, Channel pruning for accelerating very deep neural
  networks, in: IEEE International Conference on Computer Vision (ICCV),
  Vol.~2, 2017, p.~6.

\bibitem{liu2017learning}
Z.~Liu, J.~Li, Z.~Shen, G.~Huang, S.~Yan, C.~Zhang, Learning efficient
  convolutional networks through network slimming, in: IEEE International
  Conference on Computer Vision (ICCV), 2017, pp. 2755--2763.

\bibitem{li2016pruning}
H.~Li, A.~Kadav, I.~Durdanovic, H.~Samet, H.~P. Graf, Pruning filters for
  efficient convnets, in: International Conference on Learning Representations
  (ICLR), 2017.

\bibitem{vedaldi2015matconvnet}
A.~Vedaldi, K.~Lenc, Matconvnet: Convolutional neural networks for {MATLAB},
  in: ACM International Multimedia Conference (MM), 2015, pp. 689--692.

\bibitem{oymak2017near}
S.~Oymak, C.~Thrampoulidis, B.~Hassibi, Near-optimal sample complexity bounds
  for circulant binary embedding, in: IEEE International Conference on
  Acoustics, Speech and Signal Processing (ICASSP), 2017, pp. 6359--6363.

\bibitem{iccv2015_zyang_fastfoodreparameterize}
Y.~Cheng, F.~X. Yu, R.~S. Feris, S.~Kumar, A.~Choudhary, S.~F. Chang, An
  exploration of parameter redundancy in deep networks with circulant
  projections, in: IEEE International Conference on Computer Vision (ICCV),
  2015, pp. 2857--2865.

\bibitem{iccv2015_ycheng_redundancycirculantprojection}
Z.~Yang, M.~Moczulski, M.~Denil, N.~de~Freitas, A.~Smola, L.~Song, Z.~Wang,
  Deep fried {ConvNets}, in: IEEE International Conference on Computer Vision
  (ICCV), 2015, pp. 1476--1483.

\bibitem{ding2017c}
C.~Ding, S.~Liao, Y.~Wang, Z.~Li, N.~Liu, Y.~Zhuo, C.~Wang, X.~Qian, Y.~Bai,
  G.~Yuan, et~al., {CirCNN}: accelerating and compressing deep neural networks
  using block-circulant weight matrices, in: IEEE/ACM International Symposium
  on Microarchitecture (MICRO), 2017, pp. 395--408.

\bibitem{caruana2004ensemble}
R.~Caruana, A.~Niculescu~Mizil, G.~Crew, A.~Ksikes, Ensemble selection from
  libraries of models, in: International Conference on Machine Learning (ICML),
  2004, p.~18.

\bibitem{acm2006_CBucila_modelCompres}
C.~Buciluǎ, R.~Caruana, A.~Niculescu~Mizil, Model compression, in: ACM
  International Conference on Knowledge Discovery and Data Mining (KDD), 2006,
  pp. 535--541.

\bibitem{arxiv2015_ghinton_knowdistillation}
G.~Hinton, O.~Vinyals, J.~Dean, Distilling the knowledge in a neural network,
  arXiv preprint arXiv:1503.02531.

\bibitem{iclr2015_aromeo_fitnets}
A.~Romero, N.~Ballas, S.~E. Kahou, A.~Chassang, C.~Gatta, Y.~Bengio, Fitnets:
  Hints for thin deep nets, in: International Conference on Learning
  Representations (ICLR), 2015.

\bibitem{ijcnn1990_dhammerstrom_fixedarithmeticadequate}
D.~Hammerstrom, A {VLSI} architecture for high-performance, low-cost, on-chip
  learning, in: International Joint Conference on Neural Networks (IJCNN),
  IEEE, 1990, pp. 537--544.

\bibitem{icml2015_sgupta_16bit}
S.~Gupta, A.~Agrawal, K.~Gopalakrishnan, P.~Narayanan, Deep learning with
  limited numerical precision., in: International Conference on Machine
  Learning (ICML), 2015, pp. 1737--1746.

\bibitem{arxiv2014_mcourbariaux_10bits}
M.~Courbariaux, Y.~Bengio, J.~P. David, Training deep neural networks with low
  precision multiplications, arXiv preprint arXiv:1412.7024.

\bibitem{NIPS2015_mcourbariaux_binaryweights}
M.~Courbariaux, Y.~Bengio, J.~P. David, Binaryconnect: Training deep neural
  networks with binary weights during propagations, in: Conference on Neural
  Information Processing Systems (NIPS), 2015, pp. 3123--3131.

\bibitem{eccv2016_mrastegari_xnor}
M.~Rastegari, V.~Ordonez, J.~Redmon, A.~Farhadi, {XNOR-Net}: Imagenet
  classification using binary convolutional neural networks, in: European
  Conference on Computer Vision (ECCV), Springer, 2016, pp. 525--542.

\bibitem{icml2016_mkim_bitwisenn}
M.~Kim, P.~Smaragdis, Bitwise neural networks, in: International Conference on
  Machine Learning (ICML), 2016.

\bibitem{zhou2018dorefa}
S.~Zhou, Y.~Wu, Z.~Ni, X.~Zhou, H.~Wen, Y.~Zou, {DoReFa-Net}: Training low
  bitwidth convolutional neural networks with low bitwidth gradients, arXiv
  preprint arXiv:1606.06160.

\bibitem{hubara2016quantized}
I.~Hubara, M.~Courbariaux, D.~Soudry, R.~El-Yaniv, Y.~Bengio, Quantized neural
  networks: Training neural networks with low precision weights and
  activations, arXiv preprint arXiv:1609.07061.

\bibitem{kim2017kernel}
H.~Kim, J.~Sim, Y.~Choi, L.-S. Kim, A kernel decomposition architecture for
  binary-weight convolutional neural networks, in: ACM/IEEE Design Automation
  Conference (DAC), 2017, p.~60.

\bibitem{li2016ternary}
F.~Li, B.~Zhang, B.~Liu, Ternary weight networks, arXiv preprint
  arXiv:1605.04711.

\bibitem{lin2016neural}
Z.~Lin, M.~Courbariaux, R.~Memisevic, Y.~Bengio, Neural networks with few
  multiplications, in: International Conference on Learning Representations
  (ICLR), 2016.

\bibitem{alemdar2017ternary}
H.~Alemdar, V.~Leroy, A.~Prost-Boucle, F.~P{\'e}trot, Ternary neural networks
  for resource-efficient {AI} applications, in: International Joint Conference
  on Neural Networks (IJCNN), 2017, pp. 2547--2554.

\bibitem{brown2001stochastic}
B.~D. Brown, H.~C. Card, Stochastic neural computation. i. computational
  elements, IEEE Transactions on Computers 50~(9) (2001) 891--905.

\bibitem{li2017structural}
Z.~Li, A.~Ren, J.~Li, Q.~Qiu, B.~Yuan, J.~Draper, Y.~Wang, Structural design
  optimization for deep convolutional neural networks using stochastic
  computing, in: IEEE/ACM Proceedings Design, Automation and Test in Eurpoe
  (DATE), 2017, pp. 250--253.

\bibitem{kim2016dynamic}
K.~Kim, J.~Kim, J.~Yu, J.~Seo, J.~Lee, K.~Choi, Dynamic energy-accuracy
  trade-off using stochastic computing in deep neural networks, in: ACM/IEEE
  Design Automation Conference (DAC), 2016, pp. 124:1--124:6.

\bibitem{ji2015hardware}
Y.~Ji, F.~Ran, C.~Ma, D.~J. Lilja, A hardware implementation of a radial basis
  function neural network using stochastic logic, in: IEEE/ACM Proceedings
  Design, Automation and Test in Eurpoe (DATE), 2015, pp. 880--883.

\bibitem{ren2017sc}
A.~Ren, Z.~Li, C.~Ding, Q.~Qiu, Y.~Wang, J.~Li, X.~Qian, B.~Yuan, {SC-DCNN}:
  highly-scalable deep convolutional neural network using stochastic computing,
  in: ACM International Conference on Architectural Support for Programming
  Languages and Operating Systems (ASPLOS), 2017, pp. 405--418.

\bibitem{li2017normalization}
J.~Li, Z.~Yuan, Z.~Li, A.~Ren, C.~Ding, J.~Draper, S.~Nazarian, Q.~Qiu,
  B.~Yuan, Y.~Wang, Normalization and dropout for stochastic computing-based
  deep convolutional neural networks, Integration, the VLSI Journal.

\bibitem{ncaa2014_li_determinedtrain}
G.~Li, P.~Niu, X.~Duan, X.~Zhang, Fast learning network: a novel artificial
  neural network with a fast learning speed, Neural Computing and Applications
  24~(7-8) (2014) 1683--1695.

\bibitem{iccsa2013_nmohd_bpcuckoo}
N.~M. Nawi, A.~Khan, M.~Z. Rehman, A new back-propagation neural network
  optimized with cuckoo search algorithm, in: International Conference on
  Computational Science and Its Applications, Springer, 2013, pp. 413--426.

\bibitem{procenn2006_yliu_antbp}
Y.~P. Liu, M.~G. Wu, J.~X. Qian, Evolving neural networks using the hybrid of
  ant colony optimization and bp algorithms, in: International Symposium on
  Neural Networks, Springer, 2006, pp. 714--722.

\bibitem{ncaa2014_pan_gawithbp}
S.~T. Pan, M.~L. Lan, An efficient hybrid learning algorithm for neural
  network--based speech recognition systems on {FPGA} chip, Neural Computing
  and Applications 24~(7-8) (2014) 1879--1885.

\bibitem{ding2011optimizing}
S.~Ding, C.~Su, J.~Yu, An optimizing bp neural network algorithm based on
  genetic algorithm, Artificial Intelligence Review 36~(2) (2011) 153--162.

\bibitem{rumelhart1985learning}
D.~E. Rumelhart, G.~E. Hinton, R.~J. Williams, Learning internal
  representations by error propagation, Tech. rep., California Univ San Diego
  La Jolla Inst for Cognitive Science (1985).

\bibitem{jmlr2011_jduchi_adagrad}
J.~Duchi, E.~Hazan, Y.~Singer, Adaptive subgradient methods for online learning
  and stochastic optimization, Journal of Machine Learning Research 12~(Jul)
  (2011) 2121--2159.

\bibitem{arxiv2012_mzeiler_adadelta}
M.~D. Zeiler, {ADADELTA}: an adaptive learning rate method, arXiv preprint
  arXiv:1212.5701.

\bibitem{jmlr2015_dkingma_adam}
D.~Kingma, J.~Ba, Adam: A method for stochastic optimization, Journal of
  Machine Learning Research (2015) 1--13.

\bibitem{ucma2011_nhamid_bpmomentum}
N.~A. Hamid, N.~M. Nawi, R.~Ghazali, M.~N.~M. Salleh, Accelerating learning
  performance of back propagation algorithm by using adaptive gain together
  with adaptive momentum and adaptive learning rate on classification problems,
  in: International Conference on Ubiquitous Computing and Multimedia
  Applications, Springer, 2011, pp. 559--570.

\bibitem{siam2012_ynesterov_gdcoordinate}
Y.~Nesterov, Efficiency of coordinate descent methods on huge-scale
  optimization problems, SIAM Journal on Optimization 22~(2) (2012) 341--362.

\bibitem{springer2014_prichtarik_gdblockcoordinate}
P.~Richt{\'a}rik, M.~Tak{\'a}{\v{c}}, Iteration complexity of randomized
  block-coordinate descent methods for minimizing a composite function,
  Mathematical Programming 144~(1-2) (2014) 1--38.

\bibitem{cublas}
{cuBLAS}, \url{http://docs.nvidia.com/cuda/cublas}.

\bibitem{mlk}
{MLK}, \url{https://software.intel.com/en-us/intel-mkl}.

\bibitem{openblas}
{OpenBLAS}, \url{http://www.openblas.net}.

\bibitem{cho2017mec}
M.~Cho, D.~Brand, Mec: Memory-efficient convolution for deep neural network,
  in: International Conference on Machine Learning (ICML), 2017, pp. 815--824.

\bibitem{cvpr2016_alavin_fftsmallfilter}
A.~Lavin, S.~Gray, Fast algorithms for convolutional neural networks, in: IEEE
  Conference on Computer Vision and Pattern Recognition (CVPR), 2016, pp.
  4013--4021.

\bibitem{park2016zero}
H.~Park, D.~Kim, J.~Ahn, S.~Yoo, Zero and data reuse-aware fast convolution for
  deep neural networks on {GPU}, in: International Conference on
  Hardware/Software Codesign and System Synthesis, 2016, pp. 1--10.

\bibitem{xiao2017exploring}
Q.~Xiao, Y.~Liang, L.~Lu, S.~Yan, Y.-W. Tai, Exploring heterogeneous algorithms
  for accelerating deep convolutional neural networks on fpgas, in: ACM/IEEE
  Design Automation Conference (DAC), 2017, pp. 1--6.

\bibitem{procavbpa1999_benyacoub_fftfirstlayer}
S.~Ben~Yacoub, B.~Fasel, J.~Luettin, Fast face detection using {MLP} and {FFT},
  in: International Conference on Audio and Video-based Biometric Person
  Authentication, 1999, pp. 31--36.

\bibitem{arxiv2014_mmathieu_trainingffts}
M.~Mathieu, M.~Henaff, Y.~LeCun, Fast training of convolutional networks
  through ffts, arXiv preprint arXiv:1312.5851.

\bibitem{neco2015_tbrosch_frequencydomain2d3d}
T.~Brosch, R.~Tam, Efficient training of convolutional deep belief networks in
  the frequency domain for application to high-resolution {2D} and {3D} images,
  Neurocomputing.

\bibitem{ko2017design}
J.~H. Ko, B.~Mudassar, T.~Na, S.~Mukhopadhyay, Design of an energy-efficient
  accelerator for training of convolutional neural networks using
  frequency-domain computation, in: ACM/IEEE Design Automation Conference
  (DAC), 2017, pp. 1--6.

\bibitem{wu2015deep}
R.~Wu, S.~Yan, Y.~Shan, Q.~Dang, G.~Sun, Deep image: Scaling up image
  recognition, arXiv preprint arXiv:1501.02876 7~(8).

\bibitem{coates2013deep}
A.~Coates, B.~Huval, T.~Wang, D.~Wu, B.~Catanzaro, N.~Andrew, Deep learning
  with cots hpc systems, in: International Conference on Machine Learning
  (ICML), 2013, pp. 1337--1345.

\bibitem{imani2017efficient}
M.~Imani, D.~Peroni, Y.~Kim, A.~Rahimi, T.~Rosing, Efficient neural network
  acceleration on {GPGPU} using content addressable memory, in: IEEE/ACM
  Proceedings Design, Automation and Test in Eurpoe (DATE), 2017, pp.
  1026--1031.

\bibitem{izeboudjen2014new}
N.~Izeboudjen, C.~Larbes, A.~Farah, A new classification approach for neural
  networks hardware: from standards chips to embedded systems on chip,
  Artificial Intelligence Review 41~(4) (2014) 491--534.

\bibitem{iccd2013_mpeemen_memorycentricaccelerator}
M.~Peemen, A.~A. Setio, B.~Mesman, H.~Corporaal, Memory-centric accelerator
  design for convolutional neural networks, in: IEEE International Conference
  on Computer Design (ICCD), 2013, pp. 13--19.

\bibitem{procacm-sigda2015_czhang_optfpga}
C.~Zhang, P.~Li, G.~Sun, Y.~Guan, B.~Xiao, J.~Cong, Optimizing {FPGA}-based
  accelerator design for deep convolutional neural networks, in: ACM Symposium
  on FPGAs, 2015, pp. 161--170.

\bibitem{martinez2013efficient}
J.~J. Mart{\'\i}nez, J.~Garrig{\'o}s, J.~Toledo, J.~M. Ferr{\'a}ndez, An
  efficient and expandable hardware implementation of multilayer cellular
  neural networks, Neurocomputing 114 (2013) 54--62.

\bibitem{procisca2010_chakradhar_dynamicconfigurab}
S.~Chakradhar, M.~Sankaradas, V.~Jakkula, S.~Cadambi, A dynamically
  configurable coprocessor for convolutional neural networks, in: IEEE
  International Symposium on Circuits and Systems (ISCAS), 2010.

\bibitem{wang2016re}
Y.~Wang, H.~Li, X.~Li, Re-architecting the on-chip memory sub-system of
  machine-learning accelerator for embedded devices, in: IEEE/ACM International
  Conference on Computer-Aided Design (ICCAD), 2016, p.~13.

\bibitem{zhang2016caffeine}
C.~Zhang, Z.~Fang, P.~Zhou, P.~Pan, J.~Cong, Caffeine: Towards uniformed
  representation and acceleration for deep convolutional neural networks, in:
  IEEE/ACM International Conference on Computer-Aided Design (ICCAD), 2016, pp.
  1--8.

\bibitem{rahman2016efficient}
A.~Rahman, J.~Lee, K.~Choi, Efficient {FPGA} acceleration of convolutional
  neural networks using logical-{3D} compute array, in: IEEE/ACM Proceedings
  Design, Automation and Test in Eurpoe (DATE), 2016, pp. 1393--1398.

\bibitem{alwani2016fused}
M.~Alwani, H.~Chen, M.~Ferdman, P.~Milder, Fused-layer {CNN} accelerators, in:
  IEEE/ACM International Symposium on Microarchitecture (MICRO), 2016, pp.
  1--12.

\bibitem{gao2017tetris}
M.~Gao, J.~Pu, X.~Yang, M.~Horowitz, C.~Kozyrakis, Tetris: Scalable and
  efficient neural network acceleration with 3d memory, in: ACM International
  Conference on Architectural Support for Programming Languages and Operating
  Systems (ASPLOS), 2017, pp. 751--764.

\bibitem{luo2017dadiannao}
T.~Luo, S.~Liu, L.~Li, Y.~Wang, S.~Zhang, T.~Chen, Z.~Xu, O.~Temam, Y.~Chen,
  Dadiannao: A neural network supercomputer, IEEE Transactions on Computers
  66~(1) (2017) 73--88.

\bibitem{wang2017chain}
S.~Wang, D.~Zhou, X.~Han, T.~Yoshimura, {Chain-NN}: An energy-efficient {1D}
  chain architecture for accelerating deep convolutional neural networks, in:
  IEEE/ACM Proceedings Design, Automation and Test in Eurpoe (DATE), 2017, pp.
  1032--1037.

\bibitem{prociscas2010_cfarabet_hardwareparallelaccelr}
C.~Farabet, B.~Martini, P.~Akselrod, S.~Talay, Y.~LeCun, E.~Culurciello,
  Hardware accelerated convolutional neural networks for synthetic vision
  systems, in: IEEE International Symposium on Circuits and Systems (ISCAS),
  2010, pp. 257--260.

\bibitem{zhang2016cambricon}
S.~Zhang, Z.~Du, L.~Zhang, H.~Lan, S.~Liu, L.~Li, Q.~Guo, T.~Chen, Y.~Chen,
  {Cambricon-X}: An accelerator for sparse neural networks, in: IEEE/ACM
  International Symposium on Microarchitecture (MICRO), 2016, pp. 1--12.

\bibitem{kwon2018maeri}
H.~Kwon, A.~Samajdar, T.~Krishna, {MAERI}: Enabling flexible dataflow mapping
  over {DNN} accelerators via reconfigurable interconnects, in: ACM
  International Conference on Architectural Support for Programming Languages
  and Operating Systems (ASPLOS), 2018, pp. 461--475.

\bibitem{fnins2016_tgokmen_resistivedevice}
T.~Gokmen, Y.~Vlasov, Acceleration of deep neural network training with
  resistive cross-point devices: Design considerations, Frontiers in
  Neuroscience 10.

\bibitem{jackson2013nanoscale}
B.~L. Jackson, B.~Rajendran, G.~S. Corrado, M.~Breitwisch, G.~W. Burr,
  R.~Cheek, K.~Gopalakrishnan, S.~Raoux, C.~T. Rettner, A.~Padilla, et~al.,
  Nanoscale electronic synapses using phase change devices, ACM Journal on
  Emerging Technologies in Computing Systems (JETC) 9~(2) (2013) 12.

\bibitem{saighi2015plasticity}
S.~Sa{\"\i}ghi, C.~G. Mayr, T.~Serrano~Gotarredona, H.~Schmidt, G.~Lecerf,
  J.~Tomas, J.~Grollier, S.~Boyn, A.~F. Vincent, D.~Querlioz, et~al.,
  Plasticity in memristive devices for spiking neural networks, Frontiers in
  Neuroscience 9 (2015) 51.

\bibitem{seo2015chip}
J.~Seo, B.~Lin, M.~Kim, P.~Y. Chen, D.~Kadetotad, Z.~Xu, A.~Mohanty,
  S.~Vrudhula, S.~Yu, J.~Ye, et~al., On-chip sparse learning acceleration with
  {CMOS} and resistive synaptic devices, IEEE Transactions on Nanotechnology
  (TNANO) 14~(6) (2015) 969--979.

\bibitem{ncaa2016_zeng_memoristor}
X.~Zeng, S.~Wen, Z.~Zeng, T.~Huang, Design of memristor-based image convolution
  calculation in convolutional neural network, Neural Computing and
  Applications (2016) 1--6.

\bibitem{shim2016low}
Y.~Shim, A.~Sengupta, K.~Roy, Low-power approximate convolution computing unit
  with domain-wall motion based “spin-memristor” for image processing
  applications, in: ACM/IEEE Design Automation Conference (DAC), 2016, pp.
  1--6.

\bibitem{ni2016line}
L.~Ni, H.~Huang, H.~Yu, On-line machine learning accelerator on digital
  {RRAM}-crossbar, in: IEEE International Symposium on Circuits and Systems
  (ISCAS), 2016, pp. 113--116.

\bibitem{xu2014parallel}
Z.~Xu, A.~Mohanty, P.~Y. Chen, D.~Kadetotad, B.~Lin, J.~Ye, S.~Vrudhula, S.~Yu,
  J.~Seo, Y.~Cao, Parallel programming of resistive cross-point array for
  synaptic plasticity, Procedia Computer Science 41 (2014) 126--133.

\bibitem{prezioso2015training}
M.~Prezioso, F.~Merrikh~Bayat, B.~Hoskins, G.~Adam, K.~K. Likharev, D.~B.
  Strukov, Training and operation of an integrated neuromorphic network based
  on metal-oxide memristors, Nature 521~(7550) (2015) 61--64.

\bibitem{cheng2017time}
M.~Cheng, L.~Xia, Z.~Zhu, Y.~Cai, Y.~Xie, Y.~Wang, H.~Yang, {TIME}: A
  training-in-memory architecture for memristor-based deep neural networks, in:
  ACM/IEEE Design Automation Conference (DAC), 2017, pp. 1--6.

\bibitem{tvlsis2017_song_rambuffer}
L.~Song, Y.~Wang, Y.~Han, H.~Li, Y.~Cheng, X.~Li, Stt-ram buffer design for
  precision-tunable general-purpose neural network accelerator, IEEE
  Transactions on Very Large Scale Integration Systems (TVLSI) 25~(4) (2017)
  1285--1296.

\bibitem{hu2016dot}
M.~Hu, J.~P. Strachan, Z.~Li, E.~M. Grafals, N.~Davila, C.~Graves, S.~Lam,
  N.~Ge, J.~J. Yang, R.~S. Williams, Dot-product engine for neuromorphic
  computing: programming {1T1M} crossbar to accelerate matrix-vector
  multiplication, in: ACM/IEEE Design Automation Conference (DAC), 2016, p.~19.

\bibitem{xia2016switched}
L.~Xia, T.~Tang, W.~Huangfu, M.~Cheng, X.~Yin, B.~Li, Y.~Wang, H.~Yang,
  Switched by input: Power efficient structure for {RRAM}-based convolutional
  neural network, in: ACM/IEEE Design Automation Conference (DAC), 2016, pp.
  1--6.

\bibitem{ankit2017resparc}
A.~Ankit, A.~Sengupta, P.~Panda, K.~Roy, Resparc: A reconfigurable and
  energy-efficient architecture with memristive crossbars for deep spiking
  neural networks, in: ACM/IEEE Design Automation Conference (DAC), 2017, pp.
  27:1--27:6.

\bibitem{cvpr2015_bliu_sparsecnn}
B.~Liu, M.~Wang, H.~Foroosh, M.~Tappen, M.~Pensky, Sparse convolutional neural
  networks, in: IEEE Conference on Computer Vision and Pattern Recognition
  (CVPR), 2015, pp. 806--814.

\bibitem{nips2015_shan_prunconnections}
S.~Han, J.~Pool, J.~Tran, W.~Dally, Learning both weights and connections for
  efficient neural network, in: Conference on Neural Information Processing
  Systems (NIPS), 2015, pp. 1135--1143.

\bibitem{arxiv2016_mcourbariaux_binaryweightsactivations}
M.~Courbariaux, I.~Hubara, D.~Soudry, R.~El~Yaniv, Y.~Bengio, Binarized neural
  networks: Training deep neural networks with weights and activations
  constrained to+ 1 or-1, arXiv preprint arXiv:1602.02830.

\bibitem{liew2016optimized}
S.~S. Liew, M.~Khalil~Hani, R.~Bakhteri, An optimized second order stochastic
  learning algorithm for neural network training, Neurocomputing 186 (2016)
  74--89.

\bibitem{cho2014co}
H.~Cho, M.~K. An, Co-clustering algorithm: Batch, mini-batch, and online,
  International Journal of Information and Electronics Engineering 4~(5) (2014)
  340.

\bibitem{dean2012large}
J.~Dean, G.~Corrado, R.~Monga, K.~Chen, M.~Devin, M.~Mao, A.~Senior, P.~Tucker,
  K.~Yang, Q.~V. Le, et~al., Large scale distributed deep networks, in:
  Conference on Neural Information Processing Systems (NIPS), 2012, pp.
  1223--1231.

\bibitem{meng2016asynchronous}
Q.~Meng, W.~Chen, J.~Yu, T.~Wang, Z.~M. Ma, T.-Y. Liu, Asynchronous accelerated
  stochastic gradient descent, in: International Joint Conference on Artificial
  Intelligence (IJCAI), 2016.

\bibitem{nips2015_orippel_dftinpoolingcnn}
O.~Rippel, J.~Snoek, R.~P. Adams, Spectral representations for convolutional
  neural networks, in: Conference on Neural Information Processing Systems
  (NIPS), 2015, pp. 2449--2457.

\bibitem{du2017accelerator}
Z.~Du, S.~Liu, R.~Fasthuber, T.~Chen, P.~Ienne, L.~Li, T.~Luo, Q.~Guo, X.~Feng,
  Y.~Chen, et~al., An accelerator for high efficient vision processing, IEEE
  Transactions on Computer-Aided Design of Integrated Circuits and Systems
  (TCAD) 36~(2) (2017) 227--240.

\bibitem{han2016eie}
S.~Han, X.~Liu, H.~Mao, J.~Pu, A.~Pedram, M.~A. Horowitz, W.~J. Dally, Eie:
  efficient inference engine on compressed deep neural network, in: IEEE/ACM
  International Symposium on Computer Architecture (ISCA), 2016, pp. 243--254.

\bibitem{qiu2016going}
J.~Qiu, J.~Wang, S.~Yao, K.~Guo, B.~Li, E.~Zhou, J.~Yu, T.~Tang, N.~Xu,
  S.~Song, et~al., Going deeper with embedded {FPGA} platform for convolutional
  neural network, in: ACM Symposium on FPGAs, 2016, pp. 26--35.

\bibitem{moons2017energy}
B.~Moons, M.~Verhelst, An energy-efficient precision-scalable {ConvNet}
  processor in 40-nm {CMOS}, IEEE Journal Solid-State Circuits 52~(4) (2017)
  903--914.

\bibitem{chen2017eyeriss}
Y.~H. Chen, T.~Krishna, J.~S. Emer, V.~Sze, Eyeriss: An energy-efficient
  reconfigurable accelerator for deep convolutional neural networks, IEEE
  Journal Solid-State Circuits 52~(1) (2017) 127--138.

\end{thebibliography}

\end{document}